# Learning Dexterous Manipulation Policies from Experience and Imitation

**Vikash Kumar[1], Abhishek Gupta[2], Emanuel Todorov[1] and Sergey Levine[2]**

**Abstract**

We explore learning-based approaches for feedback control of a dexterous five-finger hand performing non-prehensile manipulation. First, we learn local controllers that are able to perform the task starting at a predefined initial state. These controllers are constructed using trajectory optimization with respect to locally-linear time-varying models learned directly from sensor data. In some cases, we initialize the optimizer with human demonstrations collected via teleoperation in a virtual environment. We demonstrate that such controllers can perform the task robustly, both in simulation and on the physical platform, for a limited range of initial conditions around the trained starting state. We then consider two interpolation methods for generalizing to a wider range of initial conditions: deep learning, and nearest neighbors. We find that nearest neighbors achieve higher performance. Nevertheless, the neural network has its advantages: it uses only tactile and proprioceptive feedback but no visual feedback about the object (i.e. it performs the task blind) and learns a time-invariant policy. In contrast, the nearest neighbors method switches between time-varying local controllers based on the proximity of initial object states sensed via motion capture. While both generalization methods leave room for improvement, our work shows that (i) local trajectory-based controllers for complex non-prehensile manipulation tasks can be constructed from surprisingly small amounts of training data, and (ii) collections of such controllers can be interpolated to form more global controllers. Results are summarized in the supplementary video: https://youtu.be/E0wmO6deqjo

## 1 Introduction

Dexterous manipulation is among the most challenging control problems in robotics, and remains largely unsolved. This is due to a combination of factors including high dimensionality, intermittent contact dynamics, and under-actuation in the case of dynamic object manipulation. Here we describe our efforts to tackle this problem in a principled way. We do not rely on manually designed controllers. Instead we synthesize controllers automatically, by optimizing high-level cost functions, as well as by building off of human-provided expert demonstrations. The resulting controllers are able to manipulate freely-moving objects, as shown in Figure 1. Such non-prehensile manipulation is challenging, since the system must reason about both the kinematics and the dynamics of the interaction Lynch and Mason (1999). We present results for learning both local models and control policies that can succeed from a single initial state, as well as more generalizable global policies that can use limited onboard sensing to perform a complex grasping behavior. The small amount of data needed for learning each controller (around 60 trials on the physical hardware) indicate that the approach can practically be used to learn large repertoires of dexterous manipulation skills.

We use our ADROIT platform Kumar, Xu and Todorov (2013), which is a ShadowHand skeleton augmented with high-performance pneumatic actuators. This system has a 100-dimensional continuous state space, including the positions and velocities of 24 joints, the pressures in 40 pneumatic cylinders, and the position and velocity of the object being manipulated.

Pneumatics have non-negligible time constants (around 20 ms in our system), which is why the cylinder pressures represent additional state variables, making it difficult to apply torque-control techniques. The system also has a 40-dimensional continuous control space – namely the commands to the proportional valves regulating the flow of compressed air to the cylinders. The cylinders act on the joints through tendons. The tendons do not introduce additional state variables (since we avoid slack via pre-tensioning) but nevertheless complicate the dynamics. Overall this is a daunting system to model, let alone control.

Depending on one's preference of terminology, our method can be classified as model-based Reinforcement Learning (RL), or as adaptive optimal control (Bellman and Kalaba 1959). While RL aims to solve the same general problem as optimal control, its uniqueness comes from the emphasis on model-free learning in stochastic

[1]Department of Computer Science & Engineering, University of Washington, WA, USA
{vikash, todorov}@cs.washington.edu
[2]Department of Electrical Engineering & Computer Sciences, University of California at Berkeley, CA, USA
{abhigupta, svlevine}@berkeley.edu

**Corresponding author:**
Vikash Kumar, add postal address

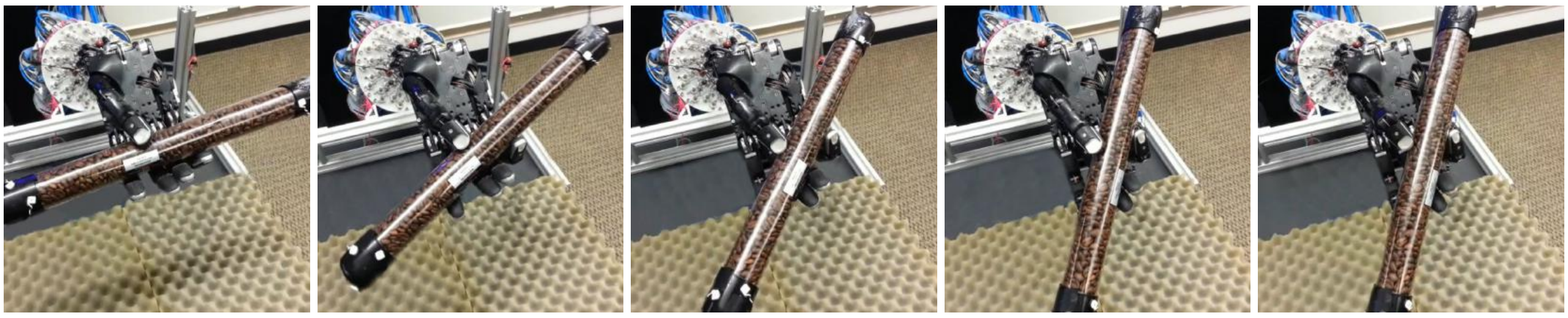

**Figure 1.** Learned hand manipulation behavior involving clockwise rotation of the object

domains (Sutton and Barto 1998). The idea of learning policies without having models still dominates RL, and forms the basis of the most remarkable success stories, both old (Tesauro 1994) and new (Mnih, Kavukcuoglu, Silver, Rusu, Veness, Bellemare, Graves, Riedmiller, Fidjeland and Ostrovski 2015). However RL with learned models has also been considered. Adaptive control on the other hand mostly focuses on learning the parameters of a model with predefined structure, essentially interleaving system identification with control (Åström and Wittenmark 2013).

Our approach here lies somewhere in between (to fix terminology, we call it RL in subsequent sections). We rely on a model, but that model does not have any informative predefined structure. Instead, it is a time-varying linear model learned from data, using a generic prior for regularization. Related ideas have been pursued previously (Mitrovic, Klanke and Vijayakumar 2010; Levine and Abbeel 2014; Levine, Wagener and Abbeel 2015b). Nevertheless, as with most approaches to automatic control and computational intelligence in general, the challenge is not only in formulating ideas but also in getting them to scale to hard problems – which is our main contribution here. In particular, we demonstrate scaling from a 14-dimensional state space in (Levine, Wagener and Abbeel 2015b) to a 100-dimensional state space here. This is important in light of the curse of dimensionality. Indeed RL has been successfully applied to a range of robotic tasks (Tedrake, Zhang and Seung 2004; Kober, Oztop and Peters 2010; Pastor, Hoffmann, Asfour and Schaal 2009; Deisenroth, Rasmussen and Fox 2011), however dimensionality and sample complexity have presented major challenges (Kober, Bagnell and Peters 2013; Deisenroth, Neumann and Peters 2013).

The manipulation skills we learn are initially represented as time-varying linear-Gaussian controllers. These controllers are fundamentally trajectory-centric, but otherwise are extremely flexible, since they can represent any trajectory with any linear stabilization strategy. Since the controllers are time-varying, the overall learned control law is nonlinear, but is locally linear at each time step. These types of controllers have been employed previously for controlling lower-dimensional robotic arms (Lioutikov, Paraschos, Neumann and Peters 2014; Levine, Wagener and Abbeel 2015b).

For learning more complex manipulation skills, we also explore the use of human demonstrations to initialize the controllers. Complex tasks with delayed rewards, such as grasping and in-hand repositioning of a heavy object, are difficult to learn from scratch. We show that a teleoperation system can be used to provide example demonstrations from a human operator using a glove-based interface, and that these demonstrations can be used to initialize learning for complex skills.

Finally, to move beyond local policies that can succeed from only a narrow range of initial states, we explore generalization through two distinct approaches. In the approach method, we train a collection of local policies, each initialized with a different initial demonstration, and then use a nearest neighbor query to select the local policy for which the initial conditions best match the current pose of the manipulated object. In the second method, we use a deep neural network to learn to mimic all of the local policies, thus removing the need for nearest neighbor queries. Our experimental results show that the nearest neighbor approach achieves the best success rate, but at the cost of requiring the variables for the nearest neighbor queries (the pose of the object) to be provided by hand. We also show that the deep neural network can learn a time-invariant policy for performing the task without requiring knowledge of the object pose at all, using only onboard sensing on the five-finger hand to perform the task.

The work on local trajectory-based control (Sections 5 and 6) was previously described in conference proceedings (Kumar, Todorov and Levine 2016) while the work on generalization (Sections 7 and 8) is new and is described here for the first time.

## 2 Related Work

Although robotic reinforcement learning has experienced considerable progress in recent years, with successful results in domains ranging from flight (Abbeel, Coates, Quigley and Ng 2006) to locomotion (Tedrake, Zhang and Seung 2004) to manipulation (Peters, Mülling and Altun 2010a; Theodorou, Buchli and Schaal 2010; Peters and Schaal 2008), comparatively few methods have been applied to control dexterous hands. (van Hoof, Hermans, Neumann and Peters 2015) report results for simple in-hand manipulation with a 3-finger hand, and our work reports learning of simple in-hand manipulation skills, such as rotating a cylinder, using time-varying linear-Gaussian controllers (Kumar, Todorov and Levine 2016). However, neither of these prior methods demonstrate generalization to conditions not seen during training. Our experiments demonstrate that our approach can learn policies for a complex precision grasping task, and can generalize to variation in the initial position of the target object. In contrast to in-hand manipulation, this task exhibits complex discontinuities at the point of contact. We overcome this challenge by combining learning from experience with imitation learning from human demonstrations, provided through a data glove teleoperation interface.

Initialization of controllers from demonstration is a widely employed technique in robotic reinforcement learning (Peters, Mülling and Altun 2010a; Theodorou, Buchli and Schaal 2010). However, most prior robotic reinforcement learning methods still use a hand-specified reward or cost function to provide the goal of the task during learning. Specifying suitable cost functions for complex dexterous manipulation can be exceedingly challenging, since simple costs can lead to poor local optima, while complex shaped costs require extensive intuition about the task. In our work, we define the cost in terms of the example demonstrations. This approach resembles the work of (Gupta, Eppner, Levine and Abbeel 2016), which used an EM-style algorithm to associate demonstrations with initial states in a reinforcement learning scenario. However, this prior work showed results on a simple inflatable hand with limited actuation, and did not demonstrate dexterous manipulation for complex tasks.

## 3 Overview

The ADROIT platform, which serves as the experimental platform for all of our dexterous manipulation experiments, is described in detail in Section 4. This system is used in three sets of experiments: the first set of experiments examines learning dexterous manipulation skills from scratch using trajectory-centric reinforcement learning, the second set of experiments is focused on learning more complex skills with a combination of trajectory-centric reinforcement learning and learning from demonstration, and the third set of experiments examines how various approximation methods, including nearest neighbor and deep neural networks, can be used to learn from multiple learned behaviors to acquire a single generalizable skill that succeeds under various circumstances.

The trajectory-centric reinforcement learning algorithm that we use combines the linear-quadratic regulator (LQR) algorithm with learned time-varying local linear models. This algorithm, which follows previous work (Levine and Abbeel 2014), is described in Section 5. We then present results on both a real-world and simulated version of the ADROIT platform using the algorithm, in Section 6. This first set of experiments focuses primarily on the capability of the trajectory-centric reinforcement learning method to efficiently learn viable and robust manipulation skills.

The second set of experiments, presented in Section 7, is aimed at evaluating how human demonstrations can be used to aid learning for more complex skills. In this section, we examine a complex grasping scenario, where trajectory-centric reinforcement learning on its own does not produce sufficiently successful behaviors, while human demonstrations alone also do not achieve a sufficient degree of robustness in the face of perturbations. We demonstrate that combining demonstrations with trajectory-centric reinforcement learning produces effective skills with a high degree of robustness to variation in the initial placement of objects in the world.

Our final set of experiments, presented in Section 8, address the question of generalization: can we use multiple skills, learned with a combination of imitation and trajectory-centric reinforcement learning, to acquire a single robust and generalizable dexterous manipulation policy? To that end, we explore the use of deep neural networks to achieve generalization by combining the behaviors of multiple skills from the previous section. We also compare to nearest neighbor as a baseline method. We demonstrate that deep neural networks can learn time-invariant manipulation policies that acquire the strategies represented by the time-varying controllers learned with trajectory-centric reinforcement learning, and furthermore can perform those skills using onboard sensing in a simulated experiment, without knowledge of the true position of the manipulated object.

## 4 System

Modularity and the ease of switching robotic platforms formed the overarching philosophy of our system design. The learning algorithm (algorithm 1) has no dependency on the selected robotic platform except for step 3, where the policies are shipped to the robotic platform for evaluation and the resulting execution trajectories are collected. This allows the training to happen either locally (on the machine controlling the robot) or remotely (if more computational power is needed).

Manipulation strategies were studied for two different platforms detailed below.

### *4.1 Hardware Platform*

The ADROIT platform is described in detail in (Kumar, Xu and Todorov 2013). Here we summarize the features relevant to the present context. ADROIT manipulation platform is an anthropomorphic arm-hand system actuated using a custom build high-performance pneumatic actuation. It consists of a 24 dof hand and a 4 dof arm. As our motivation here is to understand in-hard dexterous manipulation, we mounted the 24 dof hand on a fixed base to promote finger centric behaviors. Fixed base severely limits the workspace of the overall system but renders the system amenable to only finger-based and wrist-based manipulation strategies. For this work, we use the term 'ADROIT' to refer to the fixed base 24 dof hand setup. 20 out of the 24 hands joints are independently actuated using 40 antagonistic tendons. The DIP joints are coupled with the respective PIP joint for the 4 fingers. Finger-tendons can exert up to 42 Newton, while the wrist-tendon can exert up to 120 Newton of force. Each cylinder is supplied with compressed air via a high-performance Festo valve. The cylinders are fitted with solid-state pressure sensors. The pressures together with the joint positions and velocities (sensed by potentiometers in each joint) are provided as state variables to our controller. ADROIT's low-level driver runs on a 12 core 3.47GHz Intel(R) Xeon(R) processor with 12GB memory running Windows x64.

The manipulation task also involves an object – which is a long tube filled with coffee beans, inspired by earlier work on grasping (Amend Jr, Brown, Rodenberg, Jaeger and Lipson 2012). The object is fitted with PhaseSpace active infrared markers on each end. The markers are used to estimate the object position and velocity (both linear and angular) which are also provided as state variables. Since all our sensors have relatively low noise, we apply a minimal amount of filtering before sending the sensor data to the controller.

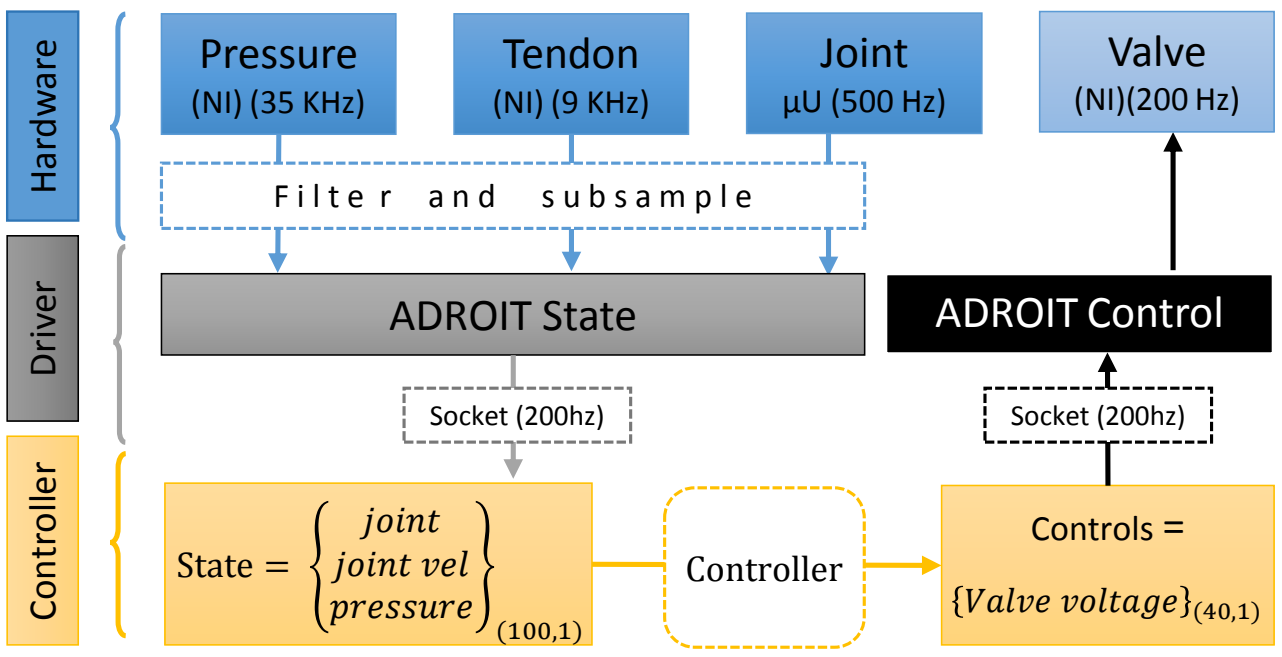

**Figure 2.** System overview

### 4.2 Simulation Platform

We model the ADROIT hand, including the antagonistic tendon transmission, joint coupling and pressure dynamics, using the MuJoCo simulator we have developed (Todorov, Erez and Tassa 2012). Pressure dynamics are implemented by extending the default actuation model with user callbacks. Simulating a 5 s trajectory at 2 ms timestep takes around 0.7 s of CPU time or around 0.3 ms of CPU time per simulation step. This includes evaluating the feedback control law (which needs to be interpolated because the trajectory optimizer uses 50 ms time steps) and advancing the physics simulation.

Having a fast simulator enables us to prototype and quickly evaluate candidate learning algorithms and cost function designs, before testing them on the hardware. Apart from being able to run much faster than real-time, the simulator can automatically reset itself to a specified initial state (which needs to be done manually on the hardware platform). Note that the actual computation time (involving GMM fitting, policy update, and network training) is practically the same for both system as the system is oblivious to the source of the observations (i.e if they were generated by the hardware or the simulation platform).

Ideally, the results on the simulation platform should be leveraged to either transfer behaviors or seed the learning on the hardware platform. This, however, is hard in practice (and still an active area of research in the field) due to (a) the difficulty in aligning the high dimensional state space of the two platforms, (b) the non-deterministic nature of the real world. State space alignment requires precise system identification and sensor calibration which otherwise are not necessary, as our algorithm can learn the local state space information directly from the raw sensor values.

## 5 Reinforcement Learning with Local Linear Models

In this section, we describe the reinforcement learning algorithm (summarised in algorithm 1) that we use to control our pneumatically-driven five finger hand. The derivation in this section follows previous work (Levine and Abbeel 2014), but we describe the algorithm in this section for completeness. The aim of the method is to learn a time-varying linear-Gaussian controller of the form $p(\mathbf{u}_t|\mathbf{x}_t) = \mathcal{N}(\mathbf{K}_t\mathbf{x}_t + \mathbf{k}_t, \mathbf{C}_t)$, where $\mathbf{x}_t$ and $\mathbf{u}_t$ are the state and action at time step $t$. The actions in our system correspond to the pneumatic valve's input voltage, while the state space is described in the preceding section. The aim of the algorithm is to minimize the expectation $E_{p(\tau)}[\ell(\tau)]$ over trajectories $\tau = \{\mathbf{x}_1, \mathbf{u}_1, \ldots, \mathbf{x}_T, \mathbf{u}_T\}$, where $\ell(\tau) = \sum_{t=1}^{T} \ell(\mathbf{x}_t, \mathbf{u}_t)$ is the total cost, and the expectation is under $p(\tau) = p(\mathbf{x}_1) \prod_{t=1}^{T} p(\mathbf{x}_{t+1}|\mathbf{x}_t, \mathbf{u}_t) p(\mathbf{u}_t|\mathbf{x}_t)$, where $p(\mathbf{x}_{t+1}|\mathbf{x}_t, \mathbf{u}_t)$ is the dynamics distribution.

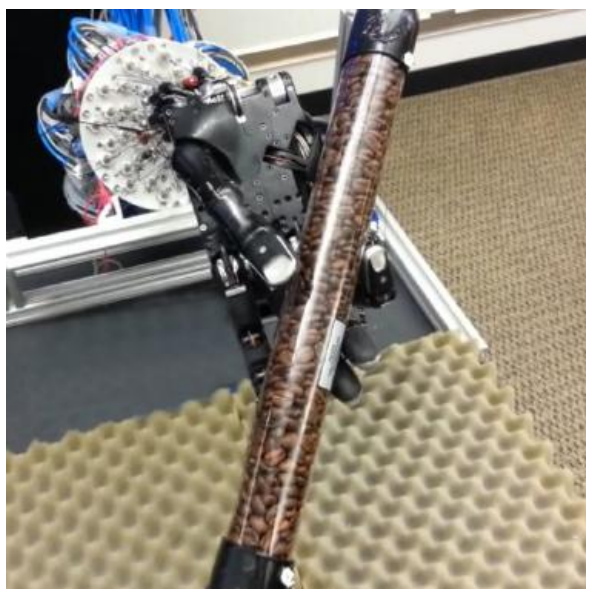

**(a)** End pose, learned

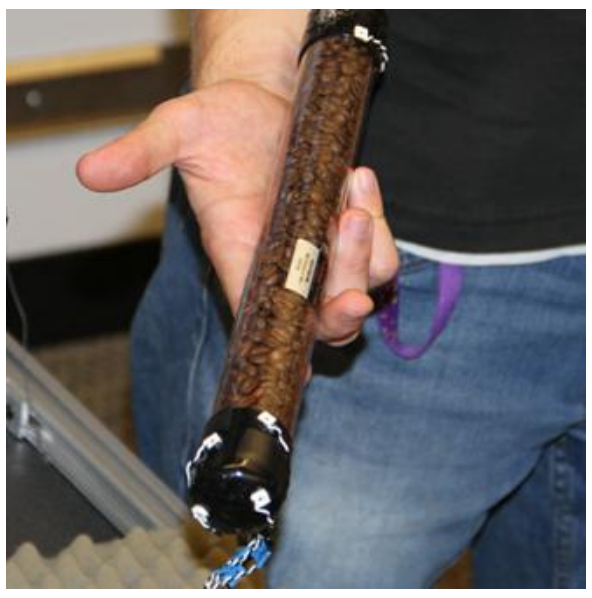

**(b)** End pose, human

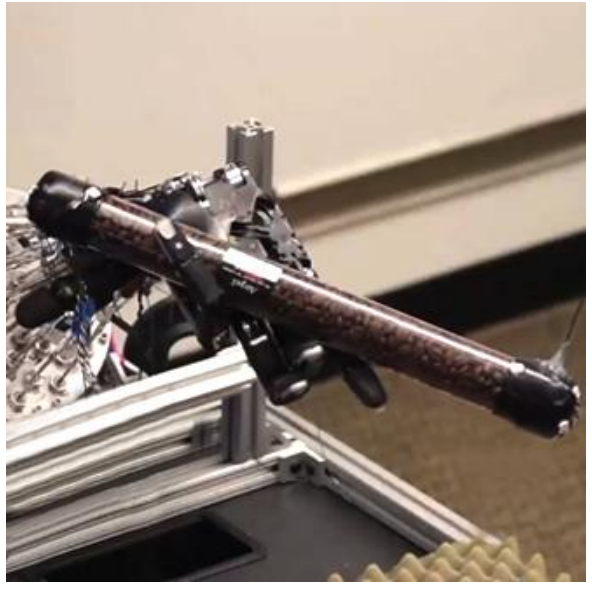

**(c)** End pose, learned

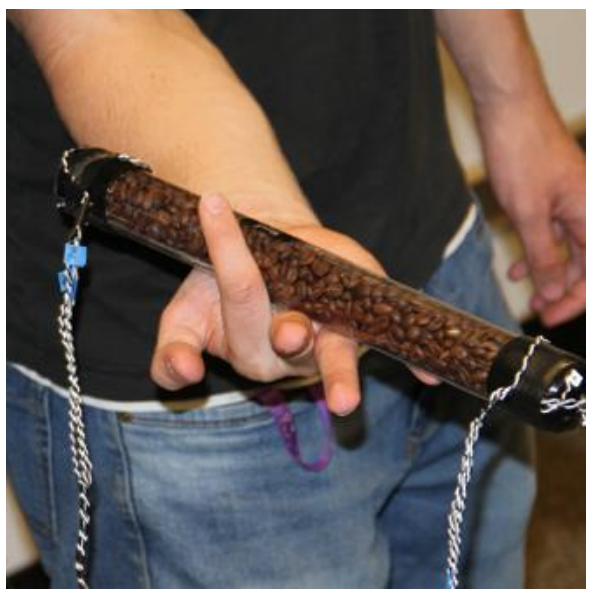

**(d)** End pose, human

**Figure 3.** Object rotation task: end poses for the two rotation directions (a-b: clockwise, c-d: anticlockwise), comparing the learned controller to the movement of a human who has not seen the robot perform the task.

**Algorithm 1** RL with linear-Gaussian controllers

1: initialize $p(\mathbf{u}_t|\mathbf{x}_t)$
2: **for** iteration $k = 1$ to $K$ **do**
3: run $p(\mathbf{u}_t|\mathbf{x}_t)$ to collect trajectory samples $\{\tau_i\}$
4: fit dynamics $p(\mathbf{x}_{t+1}|\mathbf{x}_t, \mathbf{u}_t)$ to $\{\tau_j\}$ using linear regression with GMM prior
5: fit $p = \arg\min_p E_{p(\tau)}[\ell(\tau)]$ s.t. $D_{\mathrm{KL}}(p(\tau)\|\hat{p}(\tau)) \leq \epsilon$
6: **end for**

### 5.1 Optimizing Linear-Gaussian Controllers

The simple structure of time-varying linear-Gaussian controllers admits a very efficient optimization procedure that works well even under unknown dynamics. The method is summarized in Algorithm 1. At each iteration, we run the current controller $p(\mathbf{u}_t|\mathbf{x}_t)$ on the robot to gather $N$ samples ($N = 5$ in all of our experiments), then use these samples to fit time-varying linear-Gaussian dynamics of the form $p(\mathbf{x}_{t+1}|\mathbf{x}_t, \mathbf{u}_t) = \mathcal{N}(f_{\mathbf{x}t}\mathbf{x}_t + f_{\mathbf{u}t}\mathbf{u}_t + f_{ct}, \mathbf{F}_t)$. This is done by using linear regression with a Gaussian mixture model prior, which makes it feasible to fit the dynamics even when the number of samples is much lower than the dimensionality of the system (Levine and Abbeel 2014). We also compute a second order expansion of the cost function around each of the samples, and average the expansions together to obtain a local approximate cost function of the form

$$\ell(\mathbf{x}_t, \mathbf{u}_t) \approx \frac{1}{2}[\mathbf{x}_t; \mathbf{u}_t]^{\mathrm{T}} \ell_{\mathbf{xu},\mathbf{xu}t}[\mathbf{x}_t; \mathbf{u}_t] + [\mathbf{x}_t; \mathbf{u}_t]^{\mathrm{T}} \ell_{\mathbf{xu}t} + \text{const.}$$

where subscripts denote derivatives, e.g. $\ell_{\mathbf{xu}t}$ is the gradient of $\ell$ with respect to $[\mathbf{x}_t; \mathbf{u}_t]$, while $\ell_{\mathbf{xu},\mathbf{xu}t}$ is the Hessian. The particular cost functions used in our experiments are described in the next section. When the cost function is quadratic and the dynamics are linear-Gaussian, the optimal time-varying linear-Gaussian controller of the form $p(\mathbf{u}_t|\mathbf{x}_t) = \mathcal{N}(\mathbf{K}_t\mathbf{x}_t + \mathbf{k}_t, \mathbf{C}_t)$ can be obtained by using the LQR method. This type of iterative approach can be thought of as a variant of iterative LQR (Li and Todorov 2004), where the dynamics are fitted to data. Under this model of the dynamics and cost function, the optimal policy can be computed by recursively computing the quadratic $Q$-function and value function, starting with the last time step. These functions are given by

$$V(\mathbf{x}_t) = \frac{1}{2}\mathbf{x}_t^{\mathrm{T}} V_{\mathbf{x},\mathbf{x}t}\mathbf{x}_t + \mathbf{x}_t^{\mathrm{T}} V_{\mathbf{x}t} + \text{const}$$

$$Q(\mathbf{x}_t, \mathbf{u}_t) = \frac{1}{2}[\mathbf{x}_t;\mathbf{u}_t]^{\mathrm{T}} Q_{\mathbf{xu},\mathbf{xu}t}[\mathbf{x}_t;\mathbf{u}_t] + [\mathbf{x}_t;\mathbf{u}_t]^{\mathrm{T}} Q_{\mathbf{xu}t} + \text{const}$$

We can express them with the following recurrence:

$$\begin{aligned} Q_{\mathbf{xu},\mathbf{xu}t} &= \ell_{\mathbf{xu},\mathbf{xu}t} + f_{\mathbf{xu}t}^{\mathrm{T}} V_{\mathbf{x},\mathbf{x}t+1} f_{\mathbf{xu}t} \\ Q_{\mathbf{xu}t} &= \ell_{\mathbf{xu}t} + f_{\mathbf{xu}t}^{\mathrm{T}} V_{\mathbf{x}t+1} \\ V_{\mathbf{x},\mathbf{x}t} &= Q_{\mathbf{x},\mathbf{x}t} - Q_{\mathbf{u},\mathbf{x}t}^{\mathrm{T}} Q_{\mathbf{u},\mathbf{u}t}^{-1} Q_{\mathbf{u},\mathbf{x}t} \\ V_{\mathbf{x}t} &= Q_{\mathbf{x}t} - Q_{\mathbf{u},\mathbf{x}t}^{\mathrm{T}} Q_{\mathbf{u},\mathbf{u}t}^{-1} Q_{\mathbf{u}t}, \end{aligned}$$

which allows us to compute the optimal control law as $g(\mathbf{x}_t) = \hat{\mathbf{u}}_t + \mathbf{k}_t + \mathbf{K}_t(\mathbf{x}_t - \hat{\mathbf{x}}_t)$, where $\mathbf{K}_t = -Q_{\mathbf{u},\mathbf{u}t}^{-1} Q_{\mathbf{u},\mathbf{x}t}$ and $\mathbf{k}_t = -Q_{\mathbf{u},\mathbf{u}t}^{-1} Q_{\mathbf{u}t}$. If we consider $p(\tau)$ to be the trajectory distribution formed by the deterministic control law $g(\mathbf{x}_t)$ and the stochastic dynamics $p(\mathbf{x}_{t+1}|\mathbf{x}_t, \mathbf{u}_t)$, LQR can be shown to optimize the standard objective

$$\min_{g(\mathbf{x}_t)} \sum_{t=1}^{T} E_{p(\mathbf{x}_t,\mathbf{u}_t)}[\ell(\mathbf{x}_t, \mathbf{u}_t)]. \quad (1)$$

However, we can also form a time-varying linear-Gaussian controller $p(\mathbf{u}_t|\mathbf{x}_t)$, and optimize the following objective:

$$\min_{p(\mathbf{u}_t|\mathbf{x}_t)} \sum_{t=1}^{T} E_{p(\mathbf{x}_t,\mathbf{u}_t)}[\ell(\mathbf{x}_t, \mathbf{u}_t)] - \mathcal{H}(p(\mathbf{u}_t|\mathbf{x}_t)).$$

As shown in previous work (Levine and Koltun 2013), this objective is in fact optimized by setting $p(\mathbf{u}_t|\mathbf{x}_t) = \mathcal{N}(\mathbf{K}_t\mathbf{x}_t + \mathbf{k}_t, \mathbf{C}_t)$, where $\mathbf{C}_t = Q_{\mathbf{u},\mathbf{u}t}^{-1}$. While we ultimately aim to minimize the standard controller objective in Equation (1), this maximum entropy formulation will be a useful intermediate step for a practical learning algorithm trained with fitted time-varying linear dynamics.

### 5.2 KL-Constrained Optimization

In order for this learning method to produce good results, it is important to bound the change in the controller $p(\mathbf{u}_t|\mathbf{x}_t)$ at each iteration. The standard iterative LQR method can change the controller drastically at each iteration, which can cause it to visit parts of the state space where the fitted dynamics are arbitrarily incorrect, leading to divergence. Furthermore, due to the non-deterministic nature of the real world domains, line search based methods can get misguided leading to unreliable progress.

To address these issues, we solve the following optimization problem at each iteration:

$$\min_{p(\mathbf{u}_t|\mathbf{x}_t)} E_{p(\tau)}[\ell(\tau)] \text{ s.t. } D_{\mathrm{KL}}(p(\tau)\|\hat{p}(\tau)) \leq \epsilon,$$

where $\hat{p}(\tau)$ is the trajectory distribution induced by the previous controller. Using KL-divergence constraints for controller optimization has been proposed in a number of prior works (Bagnell and Schneider 2003; Peters and Schaal 2008; Peters, Mülling and Altün 2010b). In the case of linear-Gaussian controllers, a simple modification to the LQR algorithm described above can be used to solve this constrained problem. Recall that the trajectory distributions are given by $p(\tau) = p(\mathbf{x}_1)\prod_{t=1}^{T} p(\mathbf{x}_{t+1}|\mathbf{x}_t, \mathbf{u}_t)p(\mathbf{u}_t|\mathbf{x}_t)$. Since the dynamics of the new and old trajectory distributions are assumed to be the same, the KL-divergence is given by

$$D_{\mathrm{KL}}(p(\tau)\|\hat{p}(\tau)) = \sum_{t=1}^{T} E_{p(\mathbf{x}_t,\mathbf{u}_t)}[\log \hat{p}(\mathbf{u}_t|\mathbf{x}_t)] - \mathcal{H}(p),$$

and the Lagrangian of the constrained optimization problem is given by

$$\mathcal{L}_{\text{traj}}(p, \eta) = E_p[\ell(\tau)] + \eta[D_{\mathrm{KL}}(p(\tau)\|\hat{p}(\tau)) - \epsilon] = \left[\sum_t E_{p(\mathbf{x}_t,\mathbf{u}_t)}[\ell(\mathbf{x}_t, \mathbf{u}_t) - \eta \log \hat{p}(\mathbf{u}_t|\mathbf{x}_t)]\right] - \eta\mathcal{H}(p(\tau)) - \eta\epsilon.$$

The constrained optimization can be solved with dual gradient descent (Boyd and Vandenberghe 2004), where we alternate between minimizing the Lagrangian with respect to the primal variables, which are the parameters of $p$, and taking a subgradient step on the Lagrange multiplier $\eta$. The optimization with respect to $p$ can be performed efficiently using the LQG algorithm, by observing that the Lagrangian is simply the expectation of a quantity that does not depend on $p$ and an entropy term. As described above, LQR can be used to solve maximum entropy control problems where the objective consists of a term that does not depend on $p$, and another term that encourages high entropy. We can convert the Lagrangian primal minimization into a problem of the form

$$\min_{p(\mathbf{u}_t|\mathbf{x}_t)} \sum_{t=1}^{T} E_{p(\mathbf{x}_t,\mathbf{u}_t)}[\tilde{\ell}(\mathbf{x}_t, \mathbf{u}_t)] - \mathcal{H}(p(\mathbf{u}_t|\mathbf{x}_t))$$

by using the cost $\tilde{\ell}(\mathbf{x}_t, \mathbf{u}_t) = \frac{1}{\eta}\ell(\mathbf{x}_t, \mathbf{u}_t) - \log \hat{p}(\mathbf{u}_t|\mathbf{x}_t)$. This objective is simply obtained by dividing the Lagrangian by $\eta$. Since there is only one dual variable, dual gradient descent typically converges very quickly, usually in under 10 iterations, and because LQR is a very efficient trajectory optimization method, the entire procedure can be implemented to run very quickly.

**Table 1.** Different hand positioning task variations learned

| Platform | # different task variations learned |
|---|---|
| Hardware | 5 (move assisted by gravity)<br>2 (move against gravity) |
| Simulated | 5 (move assisted by gravity)<br>3 (move against gravity) |

**Table 2.** Different object manipulation task variations learned

| Platform | # different task variations learned |
|---|---|
| Hardware + an elongated object (Fig:3) | 2 ({clockwise & anti-clockwise} object rotations along vertical) |
| Simulated + 4 object variations | 13 ({clockwise, anti-clockwise, clockwise then anti-clockwise} object rotation along vertical<br>8 ({clockwise, anti-clockwise} object rotation along horizontal) |

We initialize $p(\mathbf{u}_t|\mathbf{x}_t)$ with a fixed covariance $\mathbf{C}_t$ and zero mean, to produce random actuation on the first iteration. The Gaussian noise used to sample from $p(\mathbf{u}_t|\mathbf{x}_t)$ is generated in advance and smoothed with a Gaussian kernel with a standard deviation of two time steps, in order to produce more temporally coherent noise.

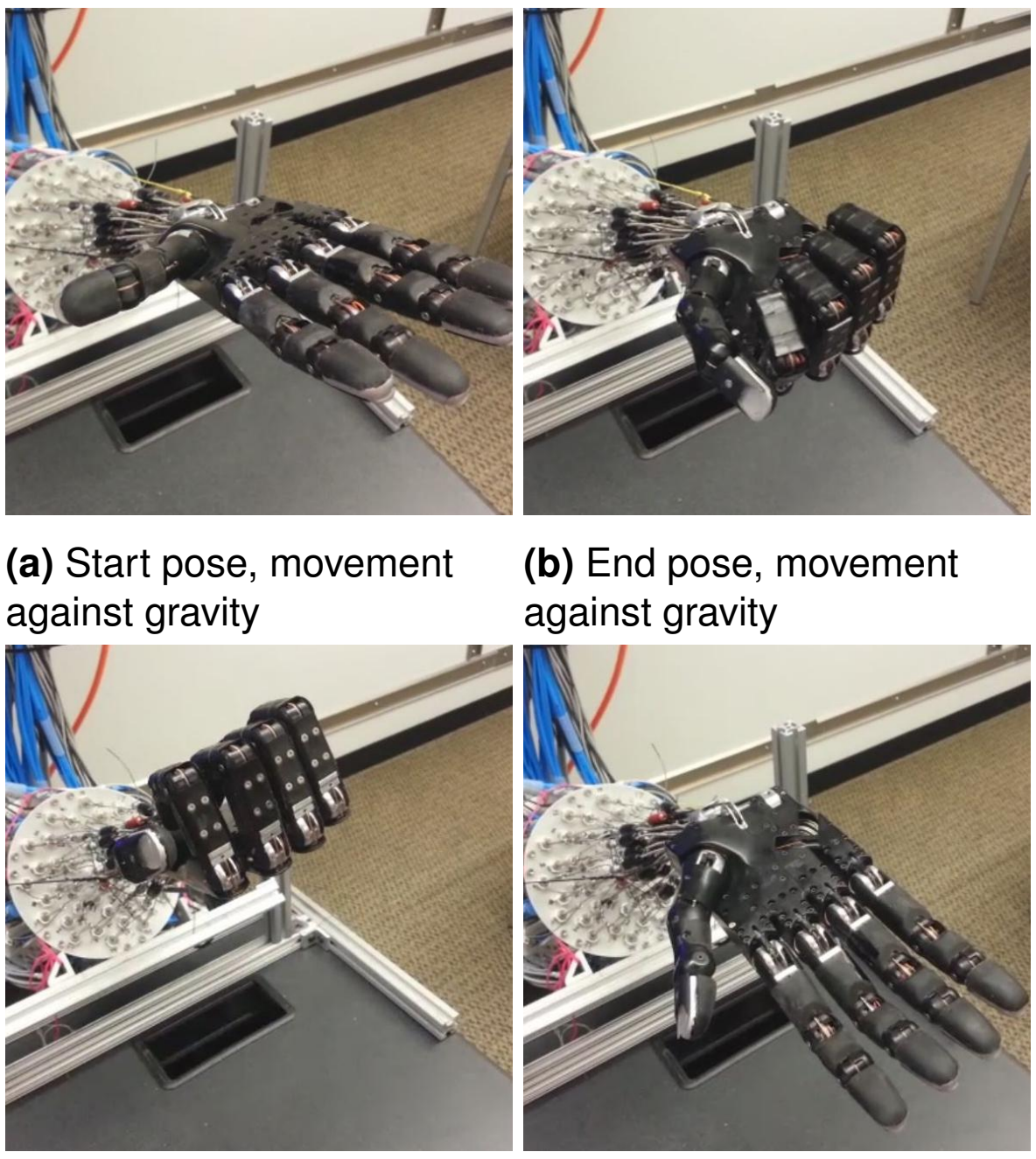

**(a)** Start pose, movement against gravity **(b)** End pose, movement against gravity **(c)** Start pose, movement assisted by gravity **(d)** End pose, movement assisted by gravity

**Figure 4.** Positioning task

# 6 Learning Policies from Experience

In this section, we will describe our first set of experiments, which uses the trajectory-centric reinforcement learning algorithm in the previous section to learn dexterous manipulation skills from scratch on both the physical and simulated ADROIT platform. The experiments in this section are aimed to ascertain whether we can learn complex manipulation behaviors entirely from scratch, using only high-level task definitions provided in terms of a cost function, with the controller learned at the level of valve opening and closing commands. The particular tasks are detailed in Table 1 and 2 and shown in the accompanying video, and include both hand posing behaviors and object manipulation skills.

## 6.1 Hand Behaviors

In the first set of tasks, we examine how well trajectory-centric reinforcement learning can control the hand to reach target poses. The state space is given by $\mathbf{x} = (q, \dot{q}, a)$. Here, $q$ denotes the vector of hand joint angles, $\dot{q}$ is the vector of joint angular velocities, $a$ the vector of cylinder pressures, and the actions $\mathbf{u}_t$ correspond to the valve command signals, which are real-valued and correspond to the degree to which each valve is opened at each time step. The tasks in this section require moving the hand to a specified pose from a given initial pose. We arranged the pair of poses such that in one task-set the finger motions were helped by gravity, and in another task-set they had to overcome gravity, as shown in Figure 4. Note that for a system of this complexity, even achieving a desired pose can be challenging, especially since the tendon actuators are in agonist-antagonist pairs and the forces have to balance to maintain posture. The cost function at each time step is given by

$$\ell(\mathbf{x}_t, \mathbf{u}_t) = ||q_t - q^*||^2 + 0.001||\mathbf{u}_t||^2,$$

and the cost at the final time step $T$ emphasizes the target pose to ensure that it is reached successfully:

$$\ell(\mathbf{x}_T, \mathbf{u}_T) = 10||q_t - q^*||^2.$$

## 6.2 Object Manipulation Behaviors

The manipulation tasks we focused on require in-hand rotation of elongated objects. We chose this task because it involves intermittent contacts with multiple fingers and is quite dynamic, while at the same time having a certain amount of intrinsic stability. We studied different variations (table 2) of this task with different objects: rotation clockwise (figure 1), rotation counter-clockwise, rotation clockwise followed by rotation counter-clockwise, and rotation clockwise without using the wrist joint (to encourage finger oriented maneuvers) – which was physically locked in that condition. Figure 3 illustrates the start and end poses and object configurations in the task learned on the ADROIT hardware platform. The running cost was

$$\begin{aligned}\ell(\mathbf{x}_t, \mathbf{u}_t) =& 0.01||q_t - q^*||^2 + 0.001||\mathbf{u}_t||^2 + \\ & ||q_t^{pos} - q^{pos*}||^2 + 10||q_t^{rot} - q^{rot*x}||^2\end{aligned}$$

where $\mathbf{x} = (q, q^{pos}, q^{rot}, \dot{q}, \dot{q}^{pos}, \dot{q}^{rot}, a)$. Here $q$ denotes the vector of hand joint angles, $q^{pos}$ the object positions, $q^{rot}$ the object rotations, $a$ the vector of cylinder pressures, and $\mathbf{u}_t$ the vector of valve command signals. At the final time we used

$$\begin{aligned}\ell(\mathbf{x}_t, \mathbf{u}_t)_{t=T} =& 2[0.01||q_t - q^*||^2 + ||q_t^{pos} - q^{pos*}||^2 \\ & + 10||q_t^{rot} - q^{rot*x}||^2].\end{aligned}$$

Here, the cost function included an extra term for desired object position and orientation. The final cost was scaled by a factor of 2 relative to the running cost.

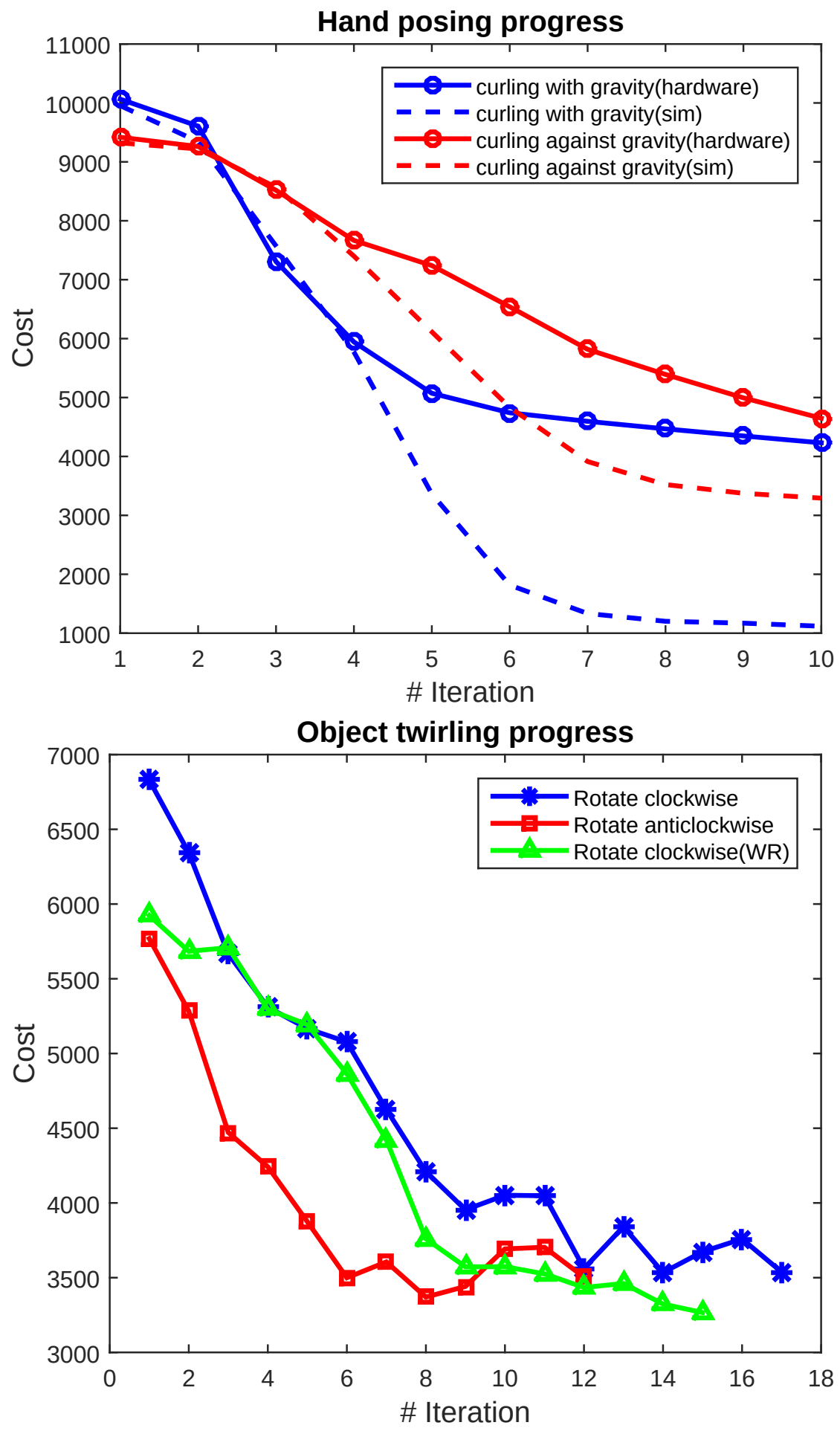

**Figure 5.** Learning curves for the positioning (top) and manipulation (bottom) tasks. *

## 6.3 Results

Besides designing the cost functions, minimal parameter tuning was required to learn each skill. Training consisted of around 15 iterations. In each iteration we performed 5 trials with different instantiations of the exploration noise in the controls. The progress of training as well as the final performance is illustrated in the video accompanying the submission, and in the figure at the beginning of the paper.

Here we quantify the performance and the robustness to noise. Figure 5 shows how the total cost for the movement (as measured by the cost functions defined above) decreased over iterations of the algorithm. The solid curves are data from the physical system. Note that in all tasks and task variations we observe very rapid convergence. Surprisingly, the manipulation task which is much harder from a control viewpoint takes about the same number of iterations to learn.

In the positioning task we also performed a systematic comparison between learning in the physical system and learning in simulation. Performance early in training was comparable, but eventually the algorithm was able to find better policies in simulation. Although it is not shown in the figure, training on simulation platform happens a lot faster, because the robot can only run in real-time while the simulated platform runs faster than real-time, and because resetting between repetitions needs to be done manually on the robot.

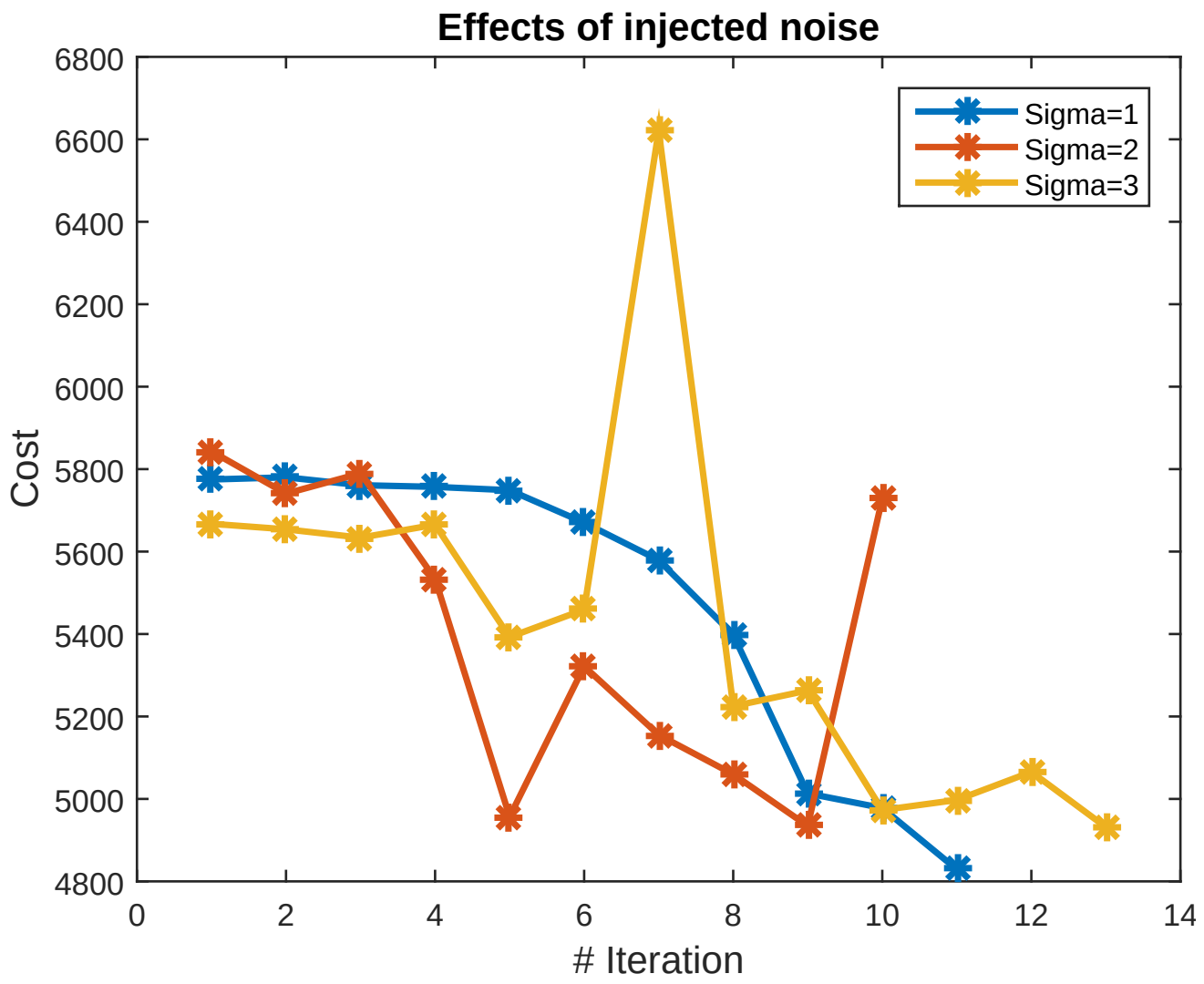

**Figure 6.** Effect of noise smoothing on learning. Sigma=1 (width of the Gaussian kernel used for smoothing noise), takes a slow start but maintains a constant progress. Higher sigma favors steep decent but it fails to maintain the progress as it is unable to successfully maintain the stability of the object being manipulated and ends up dropping it. The algorithm incurs a huge cost penalty and restarts its decent from there. *

We further investigated the effects of exploration noise magnitude injected during training. Figure 6 shows that for a relatively small amount of noise performance decreases monotonically. As we increase the noise magnitude, sometimes we see faster improvement early on but the behavior of the algorithm is no longer monotonic. These are data on the ADROIT hardware platform.

## 6.4 Delayed Robustification

Finally, we used the simulation platform to investigate robustness to perturbations more quantitatively, in the manipulation task. We wanted to quantify how robust our controllers are to changes in initial state (recall that the controllers are local). Furthermore, we wanted to see if training with noisy initial states, in addition to exploration noise injected in the controls, will result in more robust controllers. Naïvely adding initial state noise at each iteration of the algorithm (Algorithm 1) severely hindered the overall progress. However, adding initial state noise after the policy was partially learned (iteration $\geq 10$ in our case) resulted in much more robust controllers. We term this strategy as *delated robustification*.

The results of these simulations are shown in Figure 7. We plot the orientation of the object around the vertical axis as a function of time. The black curve is the unperturbed trajectory. As expected, noise injected in the initial state makes the movements more variable, especially for the controller that was trained without such noise. Adding initial state noise during training substantially improved the ability of the controller to suppress perturbations in initial state. Overall, we were surprised at how much noise we could add (up to 20 % of the range of each state variable) without the

* At each iteration, the current controller $p(\mathbf{u}_t|\mathbf{x}_t)$ is deployed on the robot to gather $N$ samples ($N = 5$ in all of our experiments).

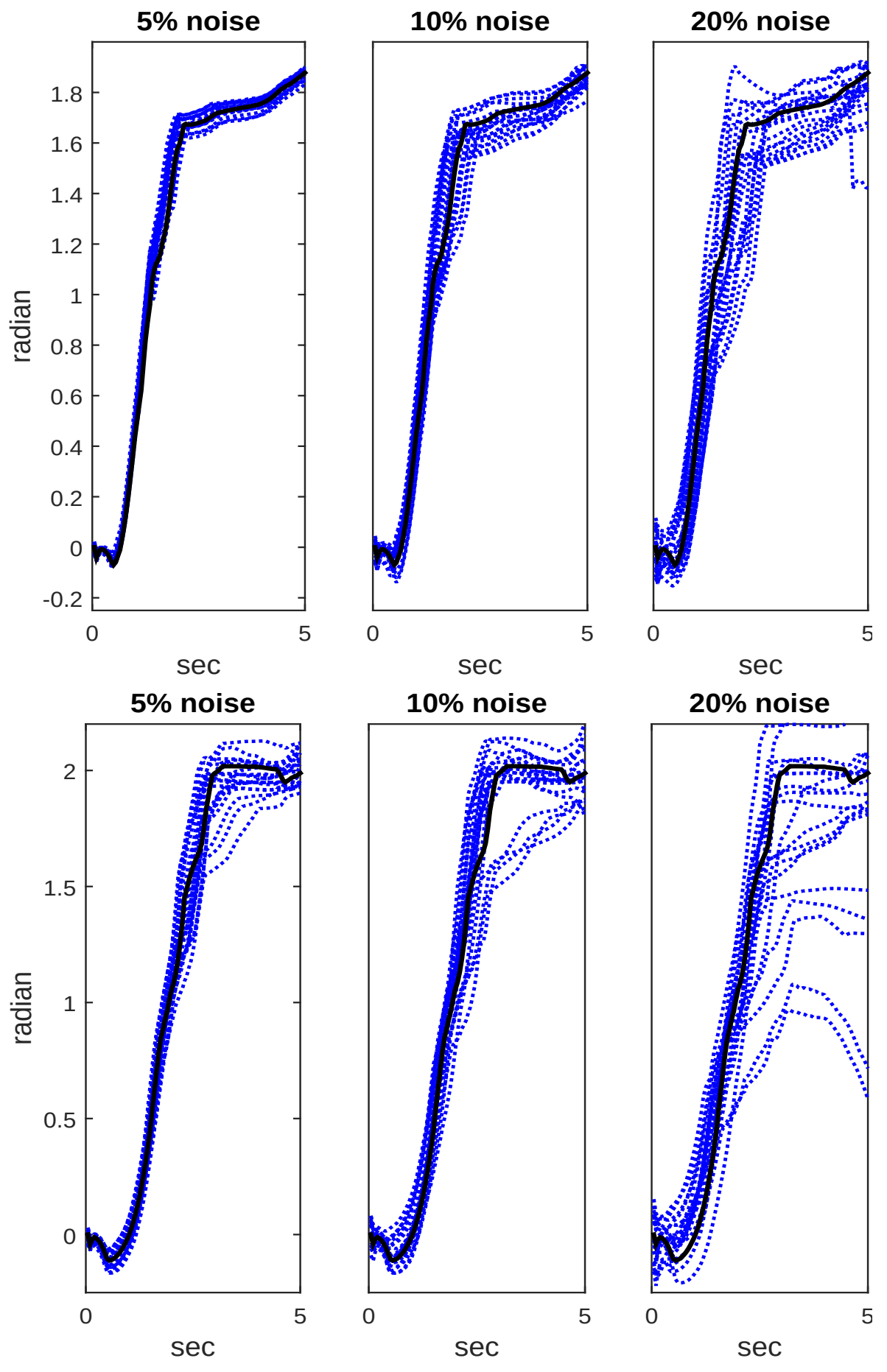

**Figure 7.** Robustness to noise in initial state. Each column corresponds to a different noise level: 5, 10, 20 % of the range of each state variable. The top row is a controller trained with noise in the initial state. The bottom row is a controller trained with the same initial state (no noise) for all trials.

hand dropping the object, in the case of the controller trained with noise. The controller trained without noise dropped the object in 4 out of 20 test trials. Thus injecting some noise in the initial state (around 2.5 %) helps improve robustness. Of course on the real robot we cannot avoid injecting such noise, because exact repositioning is very difficult.

## 7 Learning Policies from Experience and Imitation

The previous section highlighted the strengths of our reinforcement learning algorithm, outlined in Section 5, in synthesizing the details of dexterous manipulation strategies while still preserving sample efficiency. However, as our algorithm makes progress by optimizing a quadratic approximation of the cost over a local approximation of the dynamics, it can become stuck in local minima if the approximation of the learned dynamics isn't sufficiently accurate, or when the cost is not convex. In principle, arbitrary precision can be achieved by increasing the number of Gaussian kernels used for the dynamics prior, and by increasing the number of trajectory samples $N$ used for the fitting the dynamics. In practice, collecting an arbitrary number of samples might not be feasible due to time and computational limitations, especially when physical robots are involved. Furthermore, many useful task goals are non-convex, and while the method can optimize non-convex cost functions, like all local optimization methods, it does not necessarily converge to a global optimum. Random exploration can help mitigate some of these issues. To encourage exploration, we add random noise while collecting trajectory samples(step 3 of the algorithm 1). For tasks where the reward is delayed and there are multiple local minima, randomly exploring around and hoping to get lucky takes a big toll on sample efficiency. A well know strategy to overcome some of these challenges is to imitate an expert, in order to effectively steer the learned controller towards an effective solution. However, simply following an expert-provided behavior does not necessarily produce behavior that is robust to perturbations. In this section, we describe how we can combine learning from expert teleoperation with trajectory-centric reinforcement learning to acquire more complex manipulation skills that overcome local optima by following expert demonstrations while still retaining the robustness benefits of learning from experience.

### 7.1 Task Details

The task in this second set of experiments consists of picking up an elongated tube from the table. This task is challenging because the cost depends on the final configuration of the tube, which depends in a discontinuous manner on the positions of the fingers. Furthermore, grasping the tube from various initial poses involves the use of multiple different strategies. Note that the hand is rigidly mounted, so the picking must be done entirely using motion of the wrist and fingers. The shape of the tube further complicates the task. Naïve strategies like aligning the hand to the object's principal axis and force-closure will not perform well, since only the wrist is available for positioning the hand. Instead, the hand must use complex finger gaits to maneuver the object into position. The significant weight of the tube (in comparison to the lifting capability of the hand) constantly tilts the object out of the grasp. Successful strategies need to discover ways to use the thumb and fingers to act as 'support' and 'pivot' for each other, in order to reposition and reorient the object leading to a successful grasp and pick up. Failure to establish either the support or pivot results in the tube flying out of the workspace due to a net unbalanced force on the object.

The hand is mounted approximately thirty degrees to the horizontal. The task starts with the hand in zero position (Figure 8). The goal is to lift the object from a known initial condition (the object is being tracked using the PhaseSpace motion capture system). The tube is considered lifted if it is in complete control of the hand (i.e. not falling out of the hand or resting on the table) and all points on the object are above the ground by a certain height. Note that for the resemblance between the hardware and the simulated platform, the contact between the fingers and the ground plane is disabled for the simulated platform in order to allow the fingers to curl from below the object.

### 7.2 Mujoco Haptix

While leveraging an expert is quite desirable, deploying and exploiting an expert is exceptionally difficult for dexterous

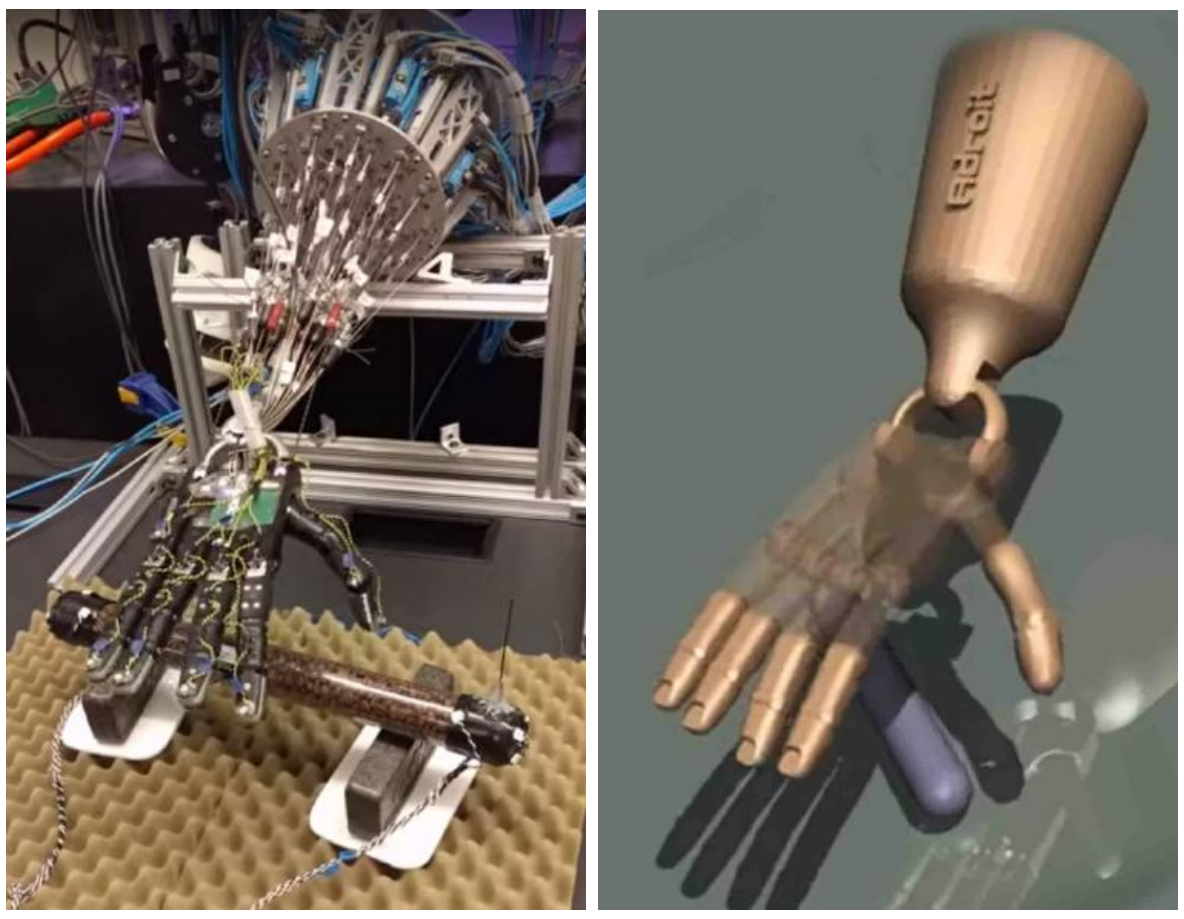

**(a)** Hardware platform **(b)** Simulation platform

**Figure 8.** Initial pose for the pickup task

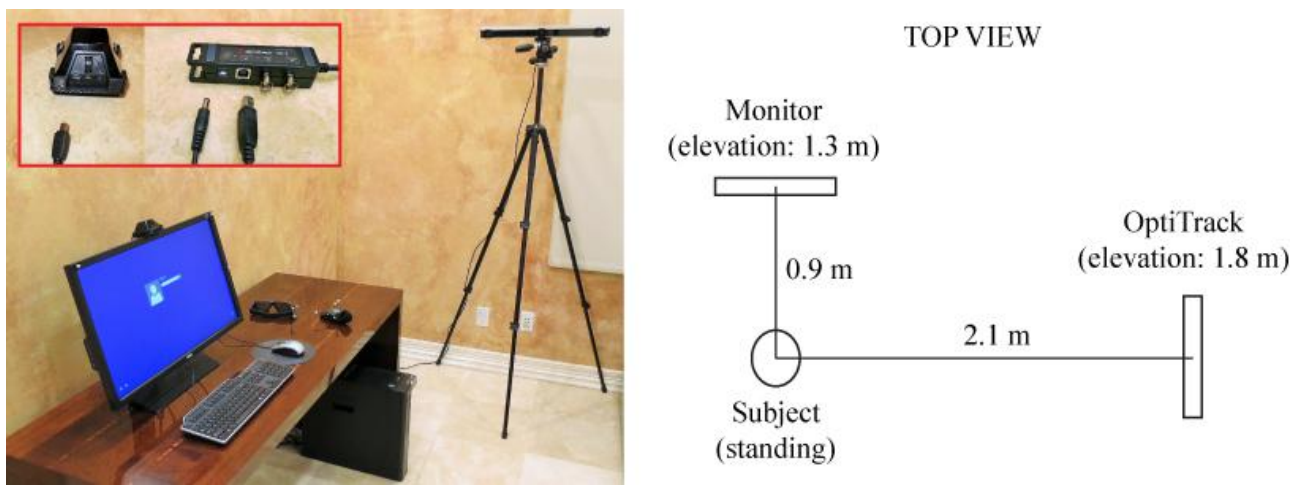

**(a)** Overall setup

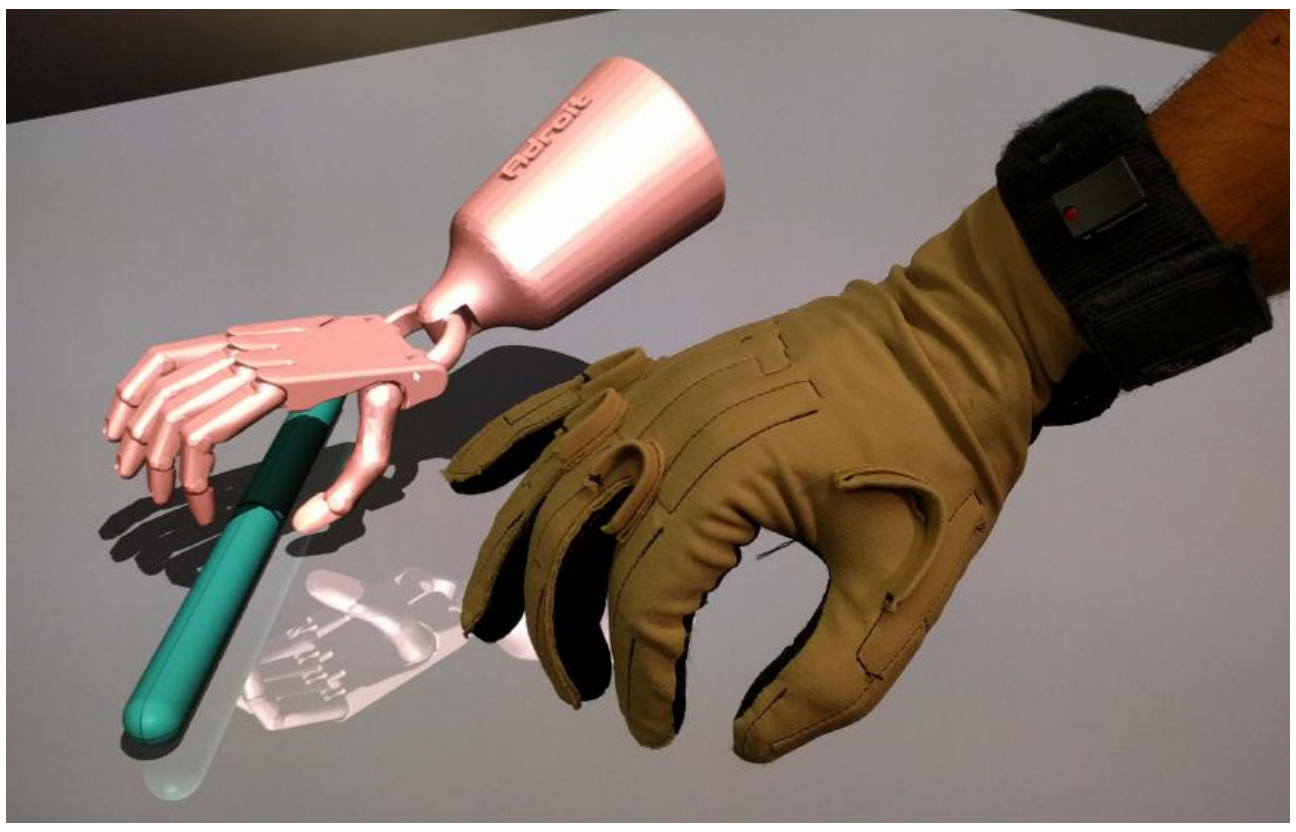

**(b)** Expert interacting with the simulation using Cyber glove system manipulation. This is due to two principal factors. First, dexterous manipulation strategies are extremely sensitive to minor variations in the contact forces, contact locations, and object positions. Thus, minor deviations from the expert demonstrations render the demonstration useless. Second, technology to capture the details of hand manipulation is often unreliable. Unlike full body movements, hand manipulation behaviors unfold in a compact region of space co-inhabited by the objects being manipulated. This makes motion capture difficult, due to occlusions and, in the case of passive systems, marker confusion. Manipulation also involves large numbers of contacts, including dynamic phenomena such as rolling, sliding, stick-slip, deformations, and soft contacts. The human hand takes advantage of these rich dynamics, but recording the data and interpreting it with regard to well-defined physics models is challenging.

**Figure 9.** Mujoco Haptix system. NVIDIA 3D Vision 2 glasses are used for stereoscopic visualization, together with a BenQ GTG XL2720Z stereo monitor. OptiTrack motion capture system is used for 3D glasses, head, and forearm tracking. A CyberGlove, calibrated for the ADROIT model, is used for tracking the finger joints.

To address these challenges, we exploit the adaptation capabilities of the brain in order to ship the data collection from the real world to a physically realistic simulation. The Mujoco Haptix system (Kumar and Todorov 2015) was developed to facilitate physically-consistent recording of rich hand-object interactions. This was done in the context of the DARPA HAPTIX program and was adapted for our purposes here. The simulation is based on the MuJoCo physics engine. The Haptix framework augments the simulator with real-time motion capture of arm and hand movements, and stereoscopic visualization using OpenGL projection from the viewpoint of the users head (which is also tracked via motion capture.) The resulting system has empirically-validated end-to-end latency of 42 msec. It creates a sense of realism which is sufficient for human users to interact with virtual objects in a natural way. Since the interaction happens in simulation, we can record every aspect of it including joint kinematics and dynamics, contact interactions, simulated sensor readings etc. There are no sensor technologies available today that could record such rich dataset from hand-object interactions in the physical world. Furthermore, since the interaction is based on our simulation model of multi-joint and contact dynamics, the dataset is by definition physically-consistent.

### 7.3 *Expert Demonstrations*

We use Mujoco Hapix to capture expert demonstrations. The expert first goes through a regression based calibration process, that maps the Cyberglove sensors to the ADROIT joint space. To enable hand manipulation, we map the joint angle $q^c$ reported by the Cyberglove to actuator commands using the Equation (2):

$$\mathbf{u}_t = \mathbf{k}J^{tendon}(\mathbf{q}_t - \mathbf{q}_t^c), \quad (2)$$

where $\mathbf{q}_t$ is the current joint configuration of the hand, $J^{tendom}$ is the tendon jacobian that maps the joint space to the tendon space, and $\mathbf{k}$ is the gain vector. $\mathbf{u}_t$ is applied as controls to the pneumatic actuators and we let the simulation evolve by stepping the physics of the world forward in time. Hand-object interactions evolve as the simulation steps forward. The expert is in a tight feedback loop with the simulation via the stereoscopic rendering. As the user interacts with the system by changing its strategy, hand manipulation behaviors emerge. We record the state $\mathbf{x}_t$ and the control trajectory $\mathbf{u}_t$ over time as demonstration trajectory. where $\mathbf{x} = (\mathbf{q}, \mathbf{q}^{pos}, \mathbf{q}^{rot}, \dot{\mathbf{q}}, \dot{\mathbf{q}}^{pos}, \dot{\mathbf{q}}^{rot}, \mathbf{a})$. Here $\mathbf{q}$ denotes the vector of hand joint angles, $\mathbf{q}^{pos}$ the object positions, $\mathbf{q}^{rot}$ the object rotations, $\mathbf{a}$ the vector of cylinder pressures, and $\mathbf{u}$ the vector of valve command signals.

### 7.4 Learning Imitation Policies

We first attempted to learn the pickup task from scratch using the following cost function:

$$\ell(\mathbf{x}_t, \mathbf{u}_t) = \alpha_1 ||q_t - q^*||^2 + \alpha_2 ||\mathbf{u}_t||^2 + \alpha_3 ||q_t^{pos} - q^{pos*}||^2 + \alpha_4 ||q_t^{rot} - q^{rot*x}||^2,$$

where $q^{pos*}$ and $q^{rot*x}$ denote the goal pose of the object, and $q^*$ is a target joint angle configuration that corresponds to a grasp. Simple learning by experience fails to come up with a strategy that can pick up the rod even after significant cost parameter tuning. The observed behavior is a combination of the following: (a) Hand randomly explores for a while and fails to find the object; (b) The fingers eventually find the object and knock it away from the manipulate-able workspace; (c) The fingers keep tapping on the top of the tube without any significant improvement. The difficulty in finding a successful pickup behavior stems from a combination of factors. First, the cost function only produces a meaningful signal toward the end of the episode, once the object is already grasped. Second, almost every strategy that can succeed has to first clear the object and only then dig the fingers into the space below the tube, with fingers on both sides to restrict the object's motion. Lastly, the high dimensionality of the ADROIT platform means that the space of successful solutions is very narrow, and a huge variety of movements are available that all fail at the task. Note that, for cases where the object isn't aligned well with the palm, additional reorientation maneuvers are required in order to pick up the object, due to the hand's restricted lateral mobility.

Our learning process for these skills combines learning from demonstrations with learning from experience, with the expert demonstration used to bootstrap the learning process. During the learning from experience phase, we use an additional shaping cost, which is a weak cost term that prevents the learning process from deviating too far away from the expert demonstration. The assumption here is that the expert demonstration is already quite good and we don't need to deviate too far to improve the solution. The strength of the shaping cost controls the amount of deviation from the expert demonstration.

For bootstrapping the learning using expert demonstrations, instead of using a random policy in step 3 of the Algorithm 1 at the initial iteration, small random noise is injected into the control trajectory of the expert demonstration to collect trajectory samples $\{\tau_i\}$. The running cost we use is

$$\ell(\mathbf{x}_t, \mathbf{u}_t) = ||\mathbf{q}_t - \hat{\mathbf{q}}_t||^2 + 0.1||\mathbf{u}_t||^2 + 50||q_t^{posZ} - 0.12||^2,$$

where $\hat{\mathbf{q}}_t$ is the hand configuration of the expert at time $t$ and $q_t^{posZ}$ is the vertical height of the object at time $t$. The first term is the shaping cost that restricts the learning from deviating too far from the demonstration, the second term is the control cost and the final term encourages picking up the object to a height of 12 cm above the ground. The final cost is the same as the running cost. Figure 10 presents a representative pickup behavior achieved using this method.

## 8 Policy Generalization

Section 6 and Section 7 explores the capabilities of trajectory-centric reinforcement learning to learn robust "local" policies for dexterous manipulation of freely-moving objects. However, the resulting local policies succeed from specific initial states, and are not designed to handle variation in the initial conditions, such as different initial placement of the manipulated object. In this section, we will discuss how we can use the local policies to learn a single global policy that generalizes effectively across different initial conditions.

Generalizable global policies can typically be learned using reinforcement learning with more expressive function approximators. Deep neural networks represent a particularly general and expressive class of function approximators, and have been recently used to learn skills that range from playing Atari games (Mnih, Kavukcuoglu, Silver, Rusu, Veness, Bellemare, Graves, Riedmiller, Fidjeland and Ostrovski 2015) to vision-based robotic manipulation (Levine, Finn, Darrell and Abbeel 2015a). In this section, we evaluate how well deep neural networks can capture generalizable policies for the pickup skill discussed in the previous section, where variation in the task corresponds to changes in the initial pose of the rod. Our goal is to explore the generalization capabilities of deep neural networks along two axes: the ability to handle variability in the initial conditions, and the ability to handle partial observability and limited sensing. To that end, we will evaluate deep neural network policies that perform the task either with full state observation, or with only the onboard sensing available on the ADROIT platform, without external motion capture markers to provide the pose of the rod.

### 8.1 Task Details

We evaluate generalization on the same pickup task as outlined in Section 7.1. To analyze generalization, we vary the orientation of the rod at the initial state. The goal is to learn a strategy for this task that succeeds for any initial rod orientation. This is particularly challenging since the robot cannot translate or reorient the wrist (since the hand is stationary), and therefore must utilize substantially different grasping strategies for different rod orientations, including the use of auxiliary finger motions to reposition the rod into the desired pose. Besides this challenge, the task also inherits all of the difficulties detailed in Section 7, including the high dimensionality of the system and the challenge of delayed rewards. To mitigate the challenge of local optima, we again use expert demonstrations. We collected a set of 10 demonstrations across 180 degrees of variation in the rod orientation. Figure 11 shows the 10 initial configurations from which an expert was engaged for providing demonstrations.

### 8.2 Local Policies

Before evaluating generalization, we first analyze the performance of the individual local policies trained with imitation and learning from experience. For each task execution, we evaluate the success or failure of trial according to the following criteria: a successful picking trial must result in the object being stationary, both extremely of the rod being above the ground by a certain height, and the entire rod aligned with the $x$-axis, so as to ensure a successful grasp into the desired goal position. Note that partially successful executions, where the tube's center of mass is lifted up but one of the end points remains on the ground, are still marked as failures. This allows us to determine not only whether the tube was picked up, but whether it was also positioned in the desired pose – an important condition for any subsequent manipulations that might be applied to it.

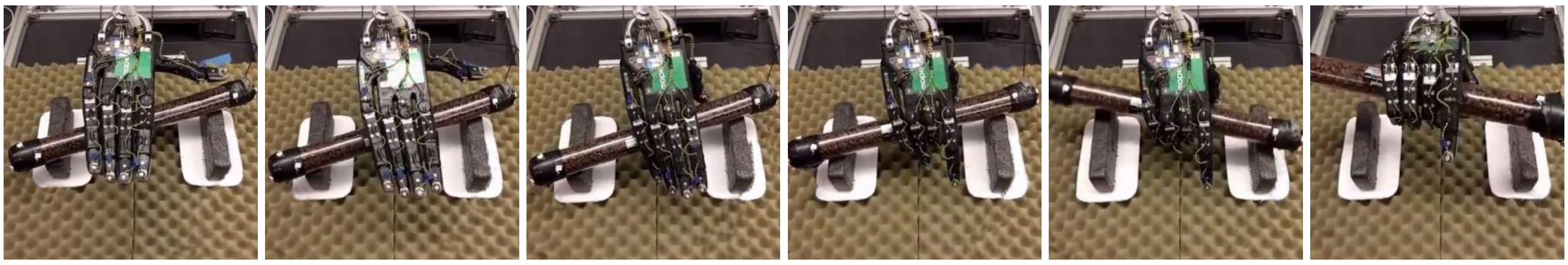

**Figure 10.** Pick up strategy learned using learning via imitation. The movement starts with aligning the palm with the object, then curling the fingers under the object followed by an object-reorientation movement that further aligns the object with the palm. The final movement engages the thumb is squeeze hard again the palm before lifting the wrist upward.

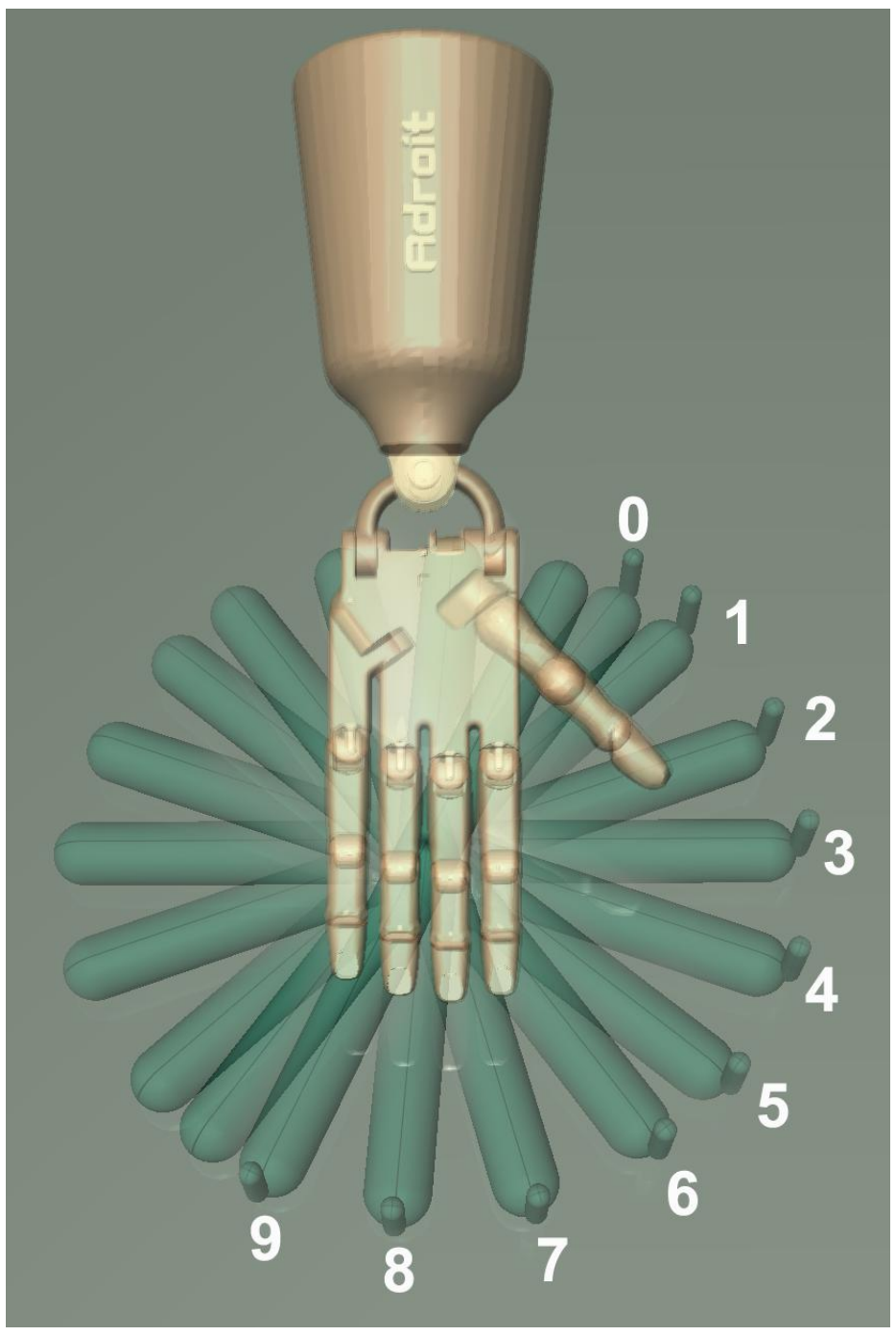

**Figure 11.** The initial tube configuration of the expert demonstration set.

In Figure 12a, we analyze the performance obtained by following the demonstrated behaviors directly, for variation in the rod angle within the local neighborhood of the rod pose for which the demonstration was generated. The numbers correspond to the conditions illustrated in Figure 11. The $x$-axis corresponds to variation in the orientation of the rod. Successful trials are marked as green circles, while failures are marked as red crosses. Overall, we observe that most demonstrations are somewhat successful for the particular rod angle for which they were created, but the success rate decreases sharply under variation in the rod pose, with discontinuous boundaries in the success region (particularly for conditions 9, 3, and 0). Some conditions, such as condition 6, are typically successful, but exhibit a brittle strategy that sometimes fails right at the demonstration angle, while others, such as conditions 3 and 1, fail sporadically at various rod poses.

As shown in Figure 12b, we can improve the robustness of the local policies corresponding to each condition by further training each of those policies from experience, following the method described in Section 7. Each demonstration was subjected to 10 iterations of learning from experience, and the resulting controllers were evaluated using the same procedure as in the previous paragraph. The controllers succeed in a wider neighborhood around their default rod pose, with the brittle strategy in condition 6 becoming much more robust, and the region of success for conditions 9, 3, and 0 expanding substantially on both sides. The overall costs and success rates for individual conditions are summarized in Figure 13. While we observe that the overall cost remained similar, the success rates for varying rod poses increased substantially, particularly for conditions 6, 8, 9, and 10. Figure 14 evaluates the effectiveness of the local policies for the entire range of task variation. We observe that the policies are less effective as they move away from the zero point (marked with dark vertical line).

### 8.3 *Nearest Neighbor*

We can observe from the results in Figure 12b that, after training, each local policy succeeds in a neighborhood that extends to the boundary of the next local policy. This suggests that a relatively simple nearest-neighbor technique could in principle allow for a nonparametric strategy to expand the success region to the entire range of rod orientations. In Figure 15, we evaluate the performance of this nearest neighbor strategy, which simply deploys the local policy trained for the rod orientation that is closest to the orientation observed in the initial state, in terms of Euclidean distance. Successful trials are marked in green and failures in red, and the overall success rate is 90.8%. Although the nearest neighbor strategy is successful in this case, it requires retaining the full set of local policies and manual specification of the variables on which the nearest neighbor queries should be performed (in this case, the rod pose). Furthermore, the nearest neighbor strategy requires us to know the rod pose, which in a physical experiment would involve the use of external motion capture markers. Can we instead learn a generalizable policy for grasping the tube that does not use any external sensing, only the onboard sensing available on the hand? In the next section, we will discuss how deep neural network function approximators can leverage the local policies to learn generalizable strategies.

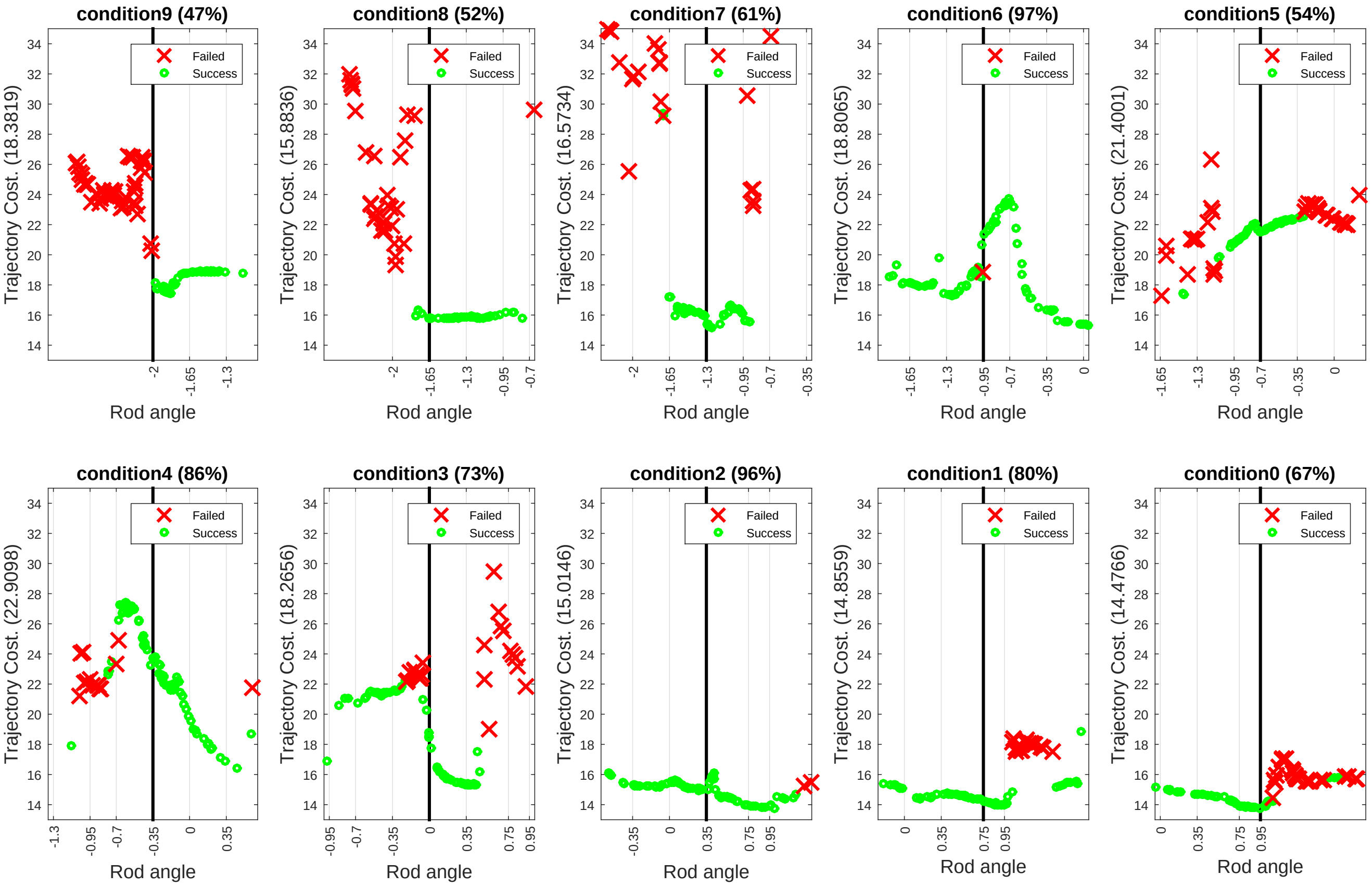

**(a)** Testing robustness of the expert demonstration in its local neighbourhood.

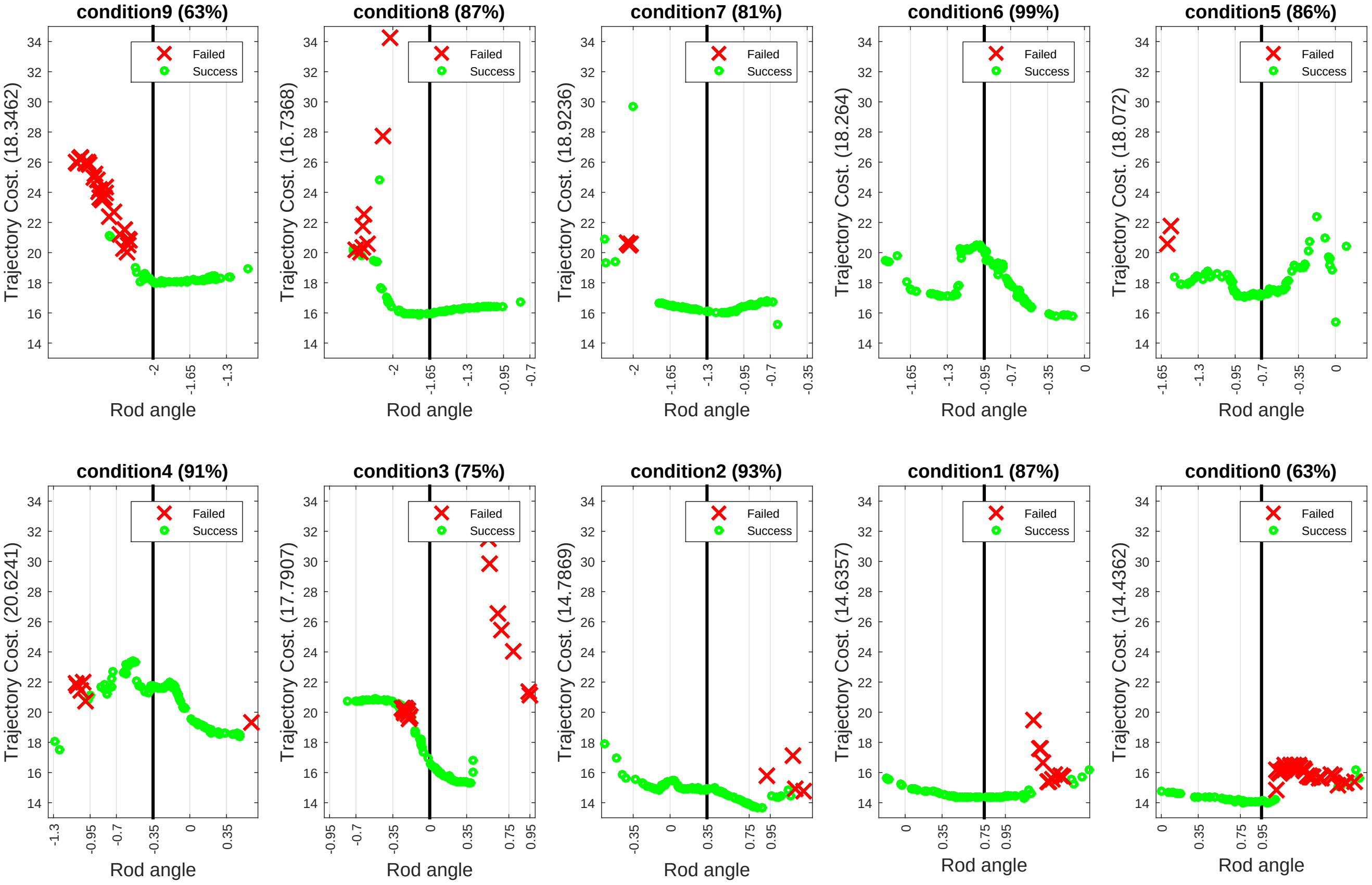

**(b)** Testing robustness of the local policies trained around the demonstrations in its local neighbourhood.

**Figure 12.** Different plots represent different conditions. Ticks on the X axis marks the different rod angles, corresponding to each condition where expert demonstration was collected. The true angle of the plot is marked with solid black line. Each condition was tested for 100 trials. Successful trials are marked in green and unsuccessful are marked in red. Success percentage are marked in the title and the average cost are provided in the Y label.

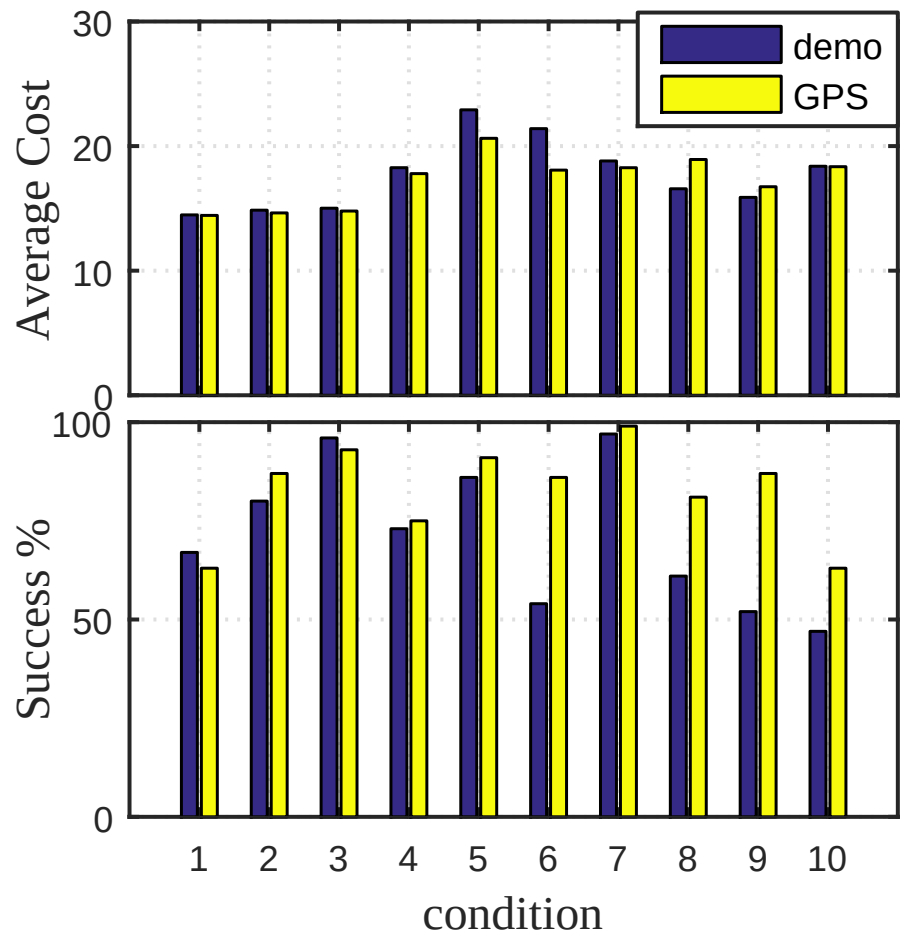

**Figure 13.** Performance comparison between the expert demonstrations and the local policies trained around the expert demonstrations in the local neighbourhood of the task (i.e. the rod angle with the vertical)

## 8.4 Generalization with Deep Neural Networks

As discussed previously, although the nearest neighbor strategy has an extremely high success rate for the pick-up task being considered, it suffers from several limitations. First, nearest neighbor methods suffer from the curse of dimensionality, which limits their applicability to high-dimensional spaces. Indeed, the nearest neighbor technique described in the preceding section requires us to manually specify that the nearest neighbor queries should be performed with respect to the pose of the rod. This information may not be readily available in general robotic manipulation tasks, where the robot must choose the strategy based on high-dimensional raw sensory information, such as camera images or inputs from tactile sensors. Second, the nearest neighbor method described above still corresponds to a time-varying control strategy, which can be limiting in cases where the robot might begin the task from arbitrary initial states. Finally, it requires storing all of the individual trajectory-centric controllers.

In this section, we evaluate whether a deep neural network representation of the rod grasping manipulation skill can be learned with comparable generalization capability and limited sensing. Deep neural networks provide a powerful and flexible class of function approximators that can be readily combined with high-dimensional inputs and complex sensing. Furthermore, the use of deep neural networks as the final policy representation does not require us to manually select a small set of state variables for nearest neighbor queries, making them a more general choice that is likely to extend to a wide variety of dexterous manipulation tasks.

To evaluate the use of deep neural networks for learning generalizable dexterous manipulation strategies, we used the collection of local policies trained from initial demonstrations to produce training data for neural network training. Each of the local policies discussed in Section 8.2 was used to generate 50 sample trajectories by executing each policy from its corresponding initial state. These samples were then used to train a deep neural network with 6 fully connected layers and 150 recitified linear (ReLU) hidden units in each layer, as shown in Figure 16. The network was trained either with the full state information provided to the local policies (which includes the pose and velocity of the rod), or with partial information reflecting onboard sensing, with knowledge about the rod pose excluded from the inputs to the neural network.

The results of the neural network policy with full state observations are shown in Figure 17. This result indicates that the neural network policy trained from the local policies is generally not as successful as the nearest neighbor strategy in the fully observed case. This is not particularly surprising: the nearest neighbor strategy already uses a set of very successful local policies that can succeed up to one angle increment, and thus form overlapping regions of effectiveness. However, the neural network must distill the distinct strategies of different local policies into a single coherent function, and therefore generally performs worse in this case. Figure 18 and Figure 19 show the performance of a large and small neural network trained without the rod pose provided as input, instead using inputs from the hand's onboard tactile sensors in the fingertips. When the network is trained without observations of the rod pose, it achieves a success rate of 74%, nearly as high as the fully observed condition, which is substantially higher than the best local policies, as shown in Figure 14.[†] These results also show that the larger network is essential for properly handling this condition. Interestingly, when the neural network is not provided with the tactile sensors either, as shown in Figure 20, it achieves nearly the same performance, indicating that the neural network is able to make use of proprioceptive sensing to determine which strategy to use. This is also not entirely surprising, since collisions and contacts result in motion of the fingers that can be detected from proprioception alone. These results indicate that the neural network policy that is only allowed to use onboard sensing learns a robust feedback behavior that can adjust the grasping strategy based on proprioceptive feedback, something that is not possible with the nearest neighbor strategy, which must choose the local policy to deploy at the beginning of the episode (before any proprioceptive sensing can take place). This suggests one of the benefits of deep neural networks for dexterous control.

## 8.5 Results on ADROIT Hardware Platform

In this section, we present our generalization results for the pick up task on the ADROIT hardware platform. The details of the task are the same as mentioned above. However, we consider generalization both in the position and orientation of the object. The orientation generalization is in a smaller neighborhood than the simulated experiments above, and we train a 6 layer fully connected neural network with 120 rectified linear (ReLU) units at each layer. The training samples are collected by sampling the local controllers learned around the 4 demonstrations provided by the expert user. The expert demonstrations were collected at the initial condition as shown in Figure 21. Local controllers were trained using 10 iterations of learning via imitation strategy as mentioned in Section 7. The training set for the network consists of 20 samples for each condition (i.e. 80 samples in total). In figure Figure 22, we cross-validate each policy for different condition, including random conditions. We found that local controllers are partially successful in a neighborhood wider than just their own. There is no local controller that works well for all the conditions. Overall the neural network policy performs at par with the local controllers on the respective conditions. The neural network policy, however, generalizes better than the individual local policies as conveyed by its higher success in picking object from random initial configurations.

[†]It is worth noting here that the hand is reset to the starting state of the trajectory controller with the closest angle for each rod pose. While these states are very similar, they may nonetheless provide additional cues to the network. Our ongoing experiments, which we will include in the final version, will address randomized initial poses to control for this potentially confounding factor.

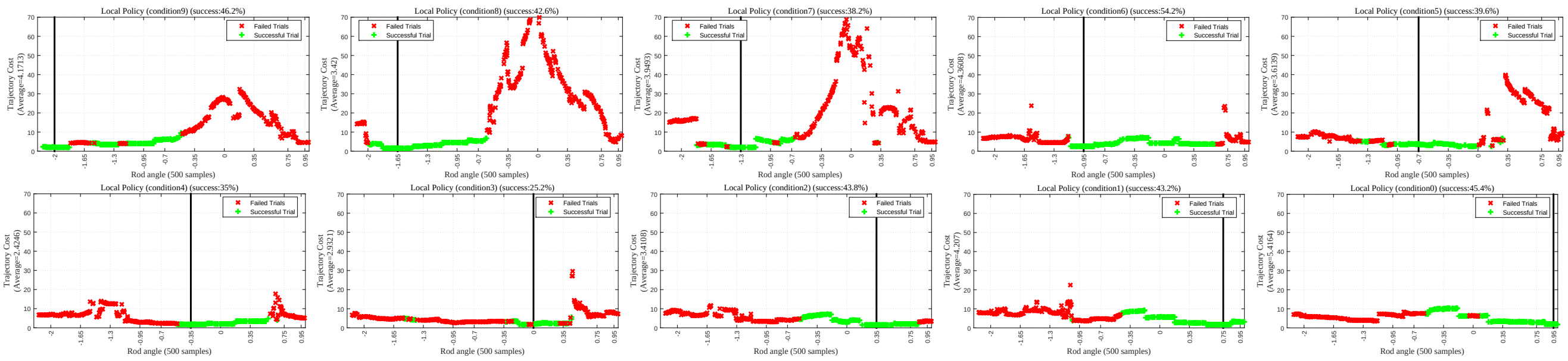

**Figure 14.** Performance of the local policies across the task variations

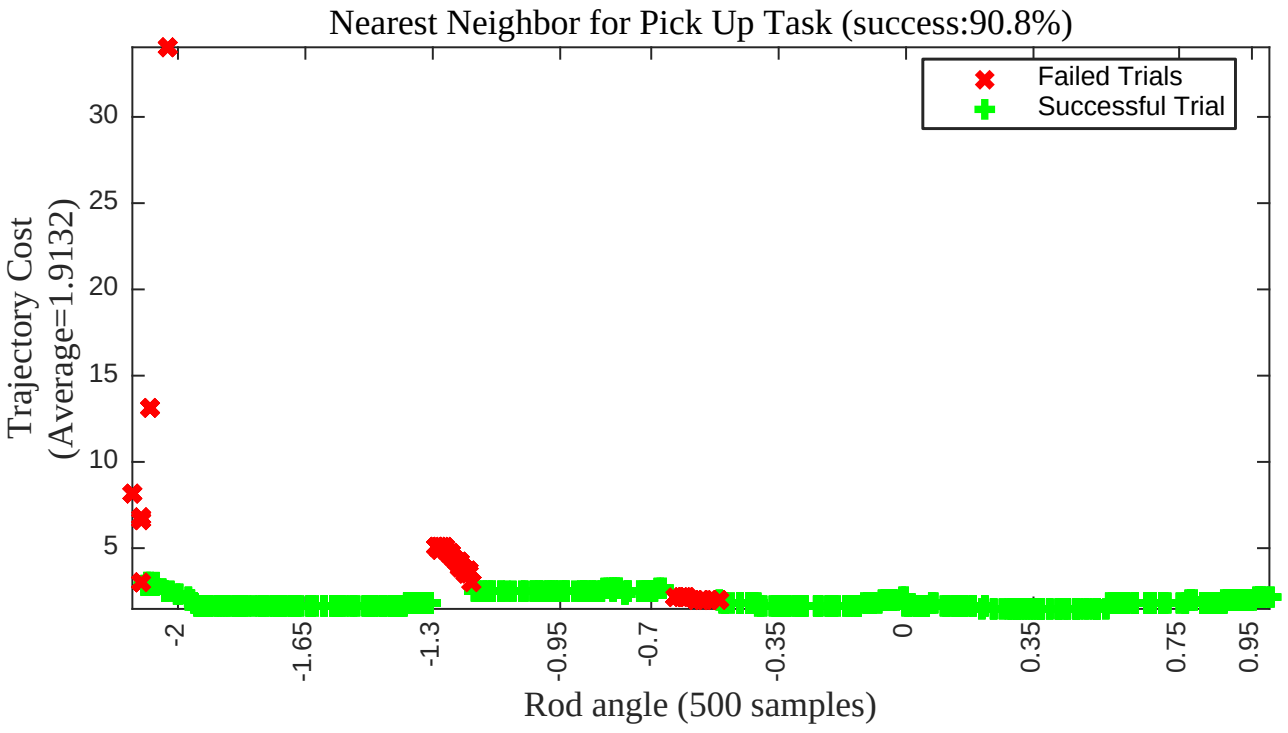

**Figure 15.** Performance of the nearest neighbor policy with full state information. Test trials are collected from random initial conditions of the object, with the local policy corresponding to the nearest rod orientation used in each trial.

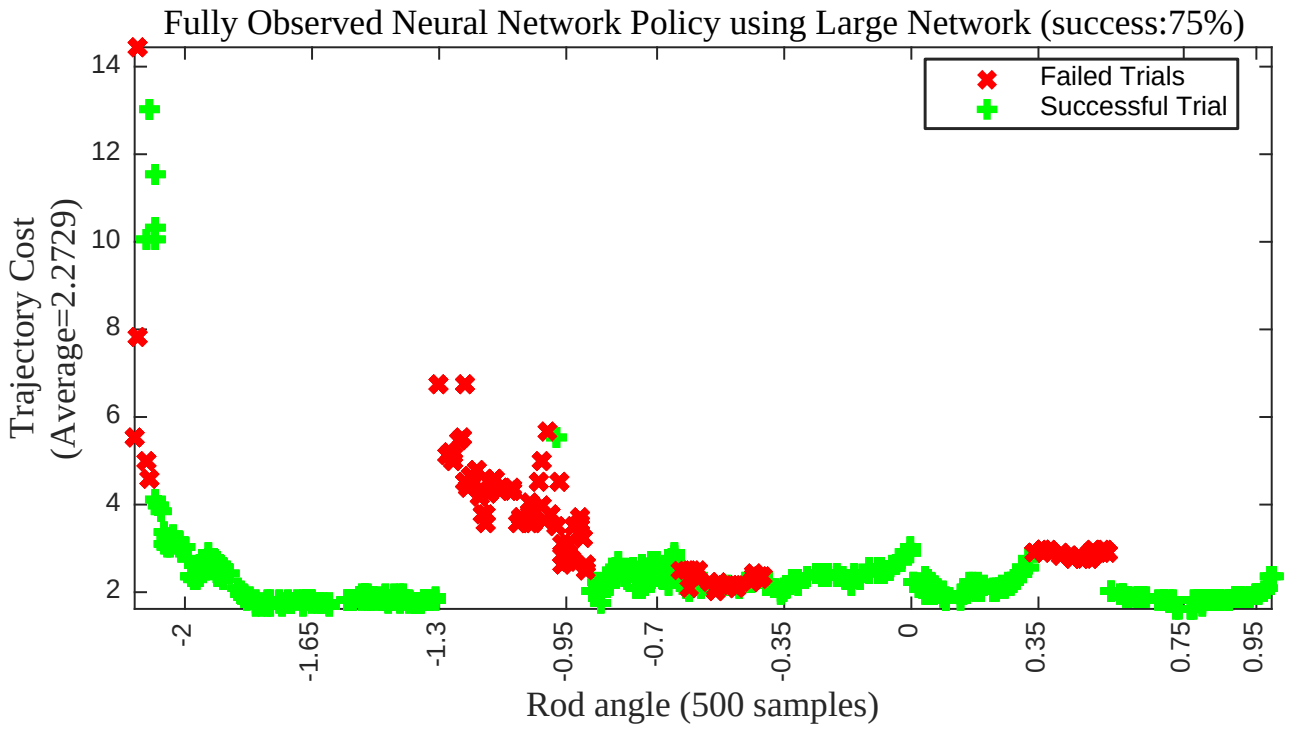

**Figure 17.** Neural network policy performance with fully observed state, at random initial object poses. The network consisted of 6 layers with 150 hidden units each, and was trained on the local policies from conditions 1, 2, 3, 4, 5, 6, and 8.

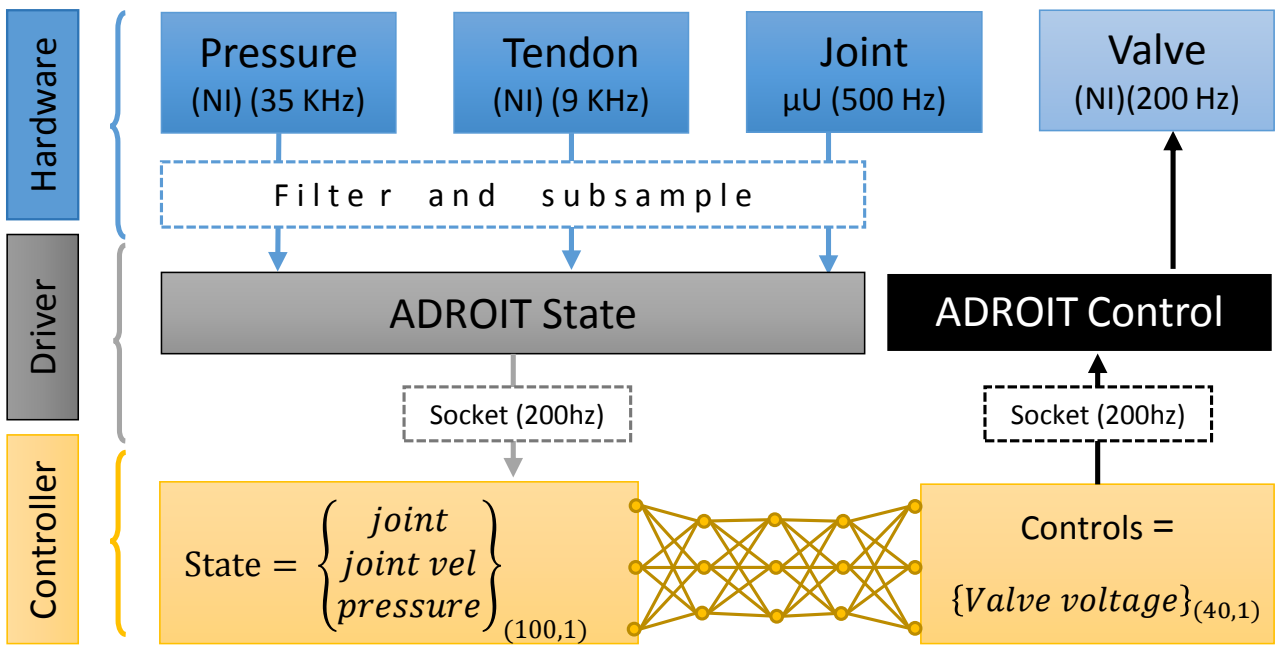

**Figure 16.** Over architecture of the system using the network as controller.

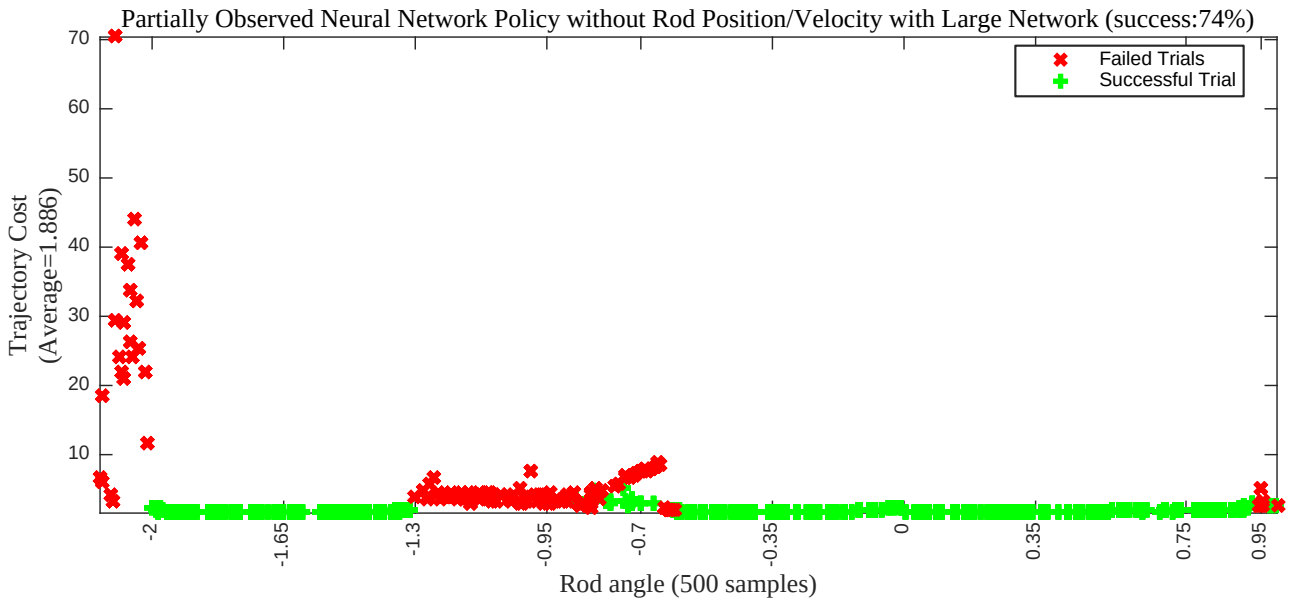

**Figure 18.** Performance of a large neural network (6 layers, 150 units) with partial observations and touch sensors. The rod position is not provided to the network, but instead the network can use inputs from the tactile sensors on the fingertips. Note that overall performance exceeds the best single local policy.

## 9 Discussion and Future Work

We demonstrated learning-based control of a complex, high-dimensional, pneumatically-driven hand. Our results include simple tasks such as reaching a target pose, as well as dynamic manipulation behaviors that involve repositioning a freely-moving cylindrical object. We also explored more complex grasping tasks that require repositioning of an object in the hand and handling delayed rewards, with the use of expert demonstrations to bootstrap learning. Aside from the high-level objective encoded in the cost function and the demonstrations, the learning algorithms do not use domain knowledge about the task or the

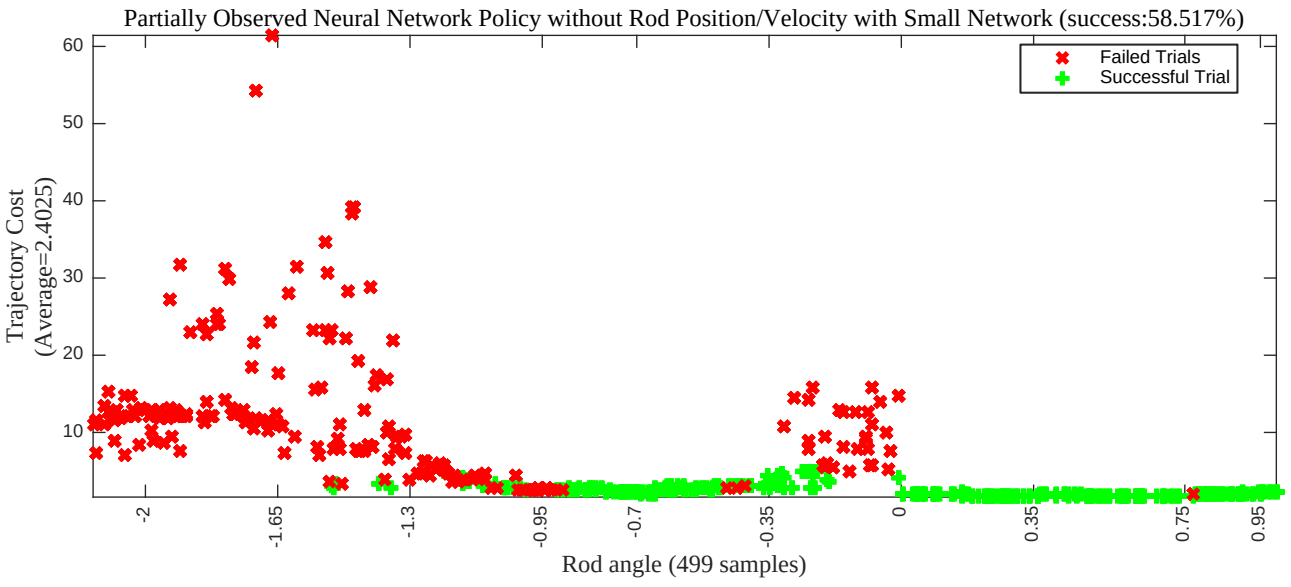

**Figure 19.** Performance of a small neural network (4 layers, 80 units) with partial observations and touch sensors. The rod position is not provided to the network, but instead the network can use inputs from the tactile sensors on the fingertips. Note that overall performance of the smaller network is substantially degraded.

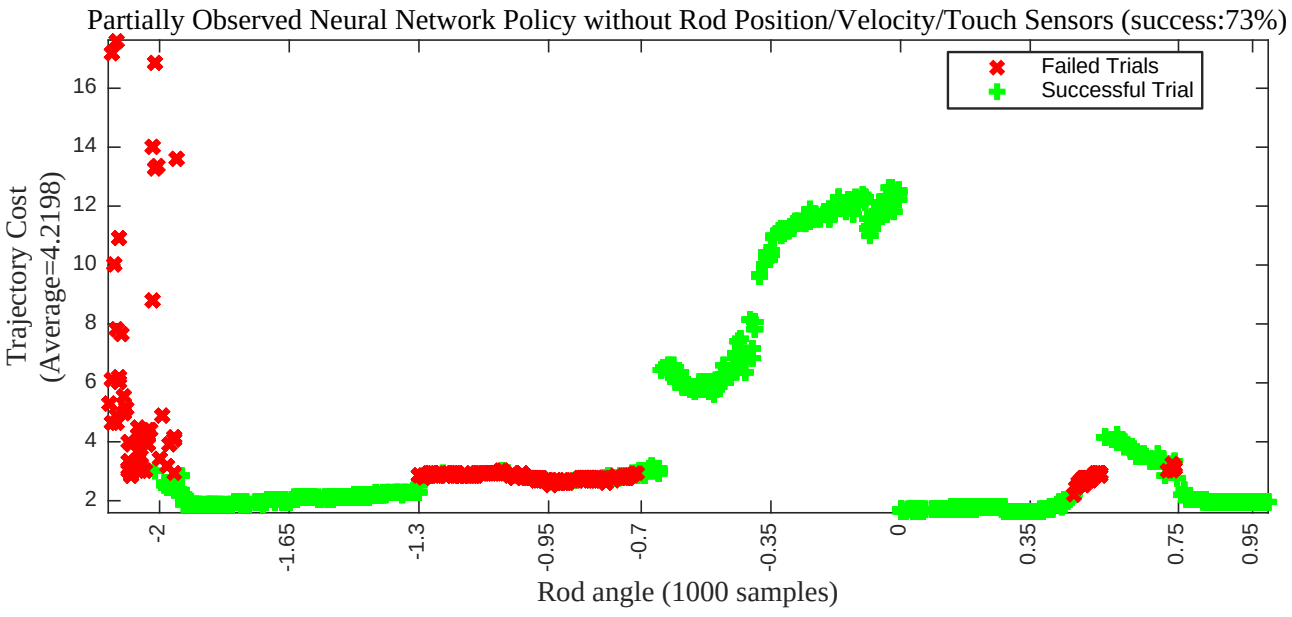

**Figure 20.** Performance of a large neural network (6 layers, 150 units) with partial observations and without touch sensors. The rod position is not provided to the network. Note that the large network performs well even without the touch sensors, indicating that the network is able to use proprioceptive inputs to determine the strategy.

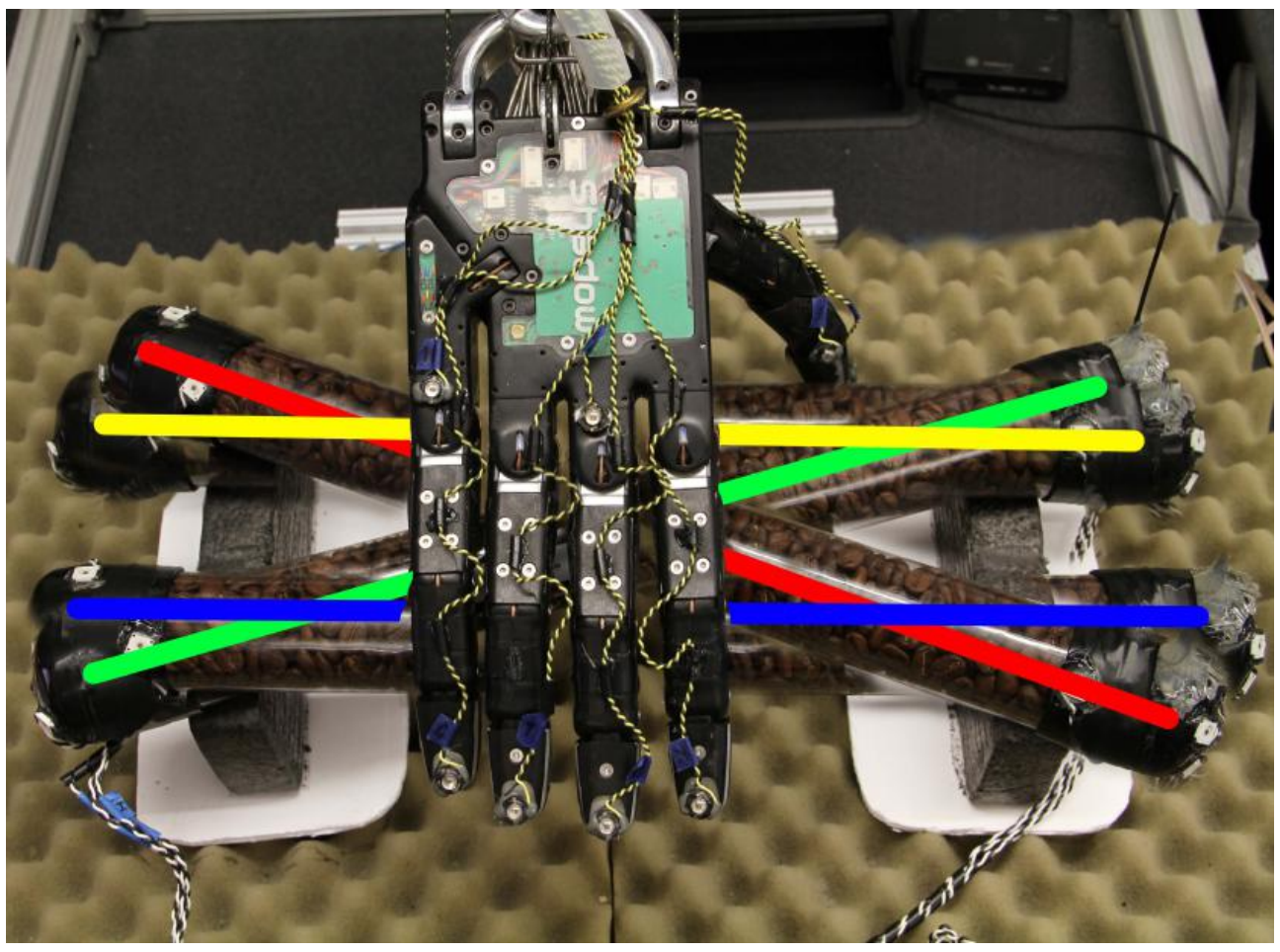

**Figure 21.** Different conditions used for expert demonstrations

hardware. The experiments show that effective manipulation strategies can be automatically discovered in this way. We further evaluated the generalization capabilities of deep neural network controllers and nearest neighbour switching controllers. While both methods were able to generalize to some extent, the performance of the nearest neighbour method was higher. The two methods have relative strengths

| | Initial condition | | | | |
|---|---|---|---|---|---|
| | A | B | C | D | Random |
| Policy-0 | 100% | 80% | 0% | 60% | 40% |
| Policy-1 | 60% | 100% | 40% | 0% | 60% |
| Policy-2 | 60% | 100% | 100% | 0% | 50% |
| Policy-3 | 100% | 40% | 0% | 100% | 60% |
| Policy-NN | 80% | 100% | 100% | 100% | 90% |
| | | | | | |
| | | 5 trials | 10 trials | | |

**Figure 22.** Cross validation of different policies under different conditions on ADROIT Hardware platform

and weaknesses and can be improved and perhaps combined in future work, as described below.

The neural network controller was given access to proprioceptive and tactile sensory input, but not to vision input about the object state. Thus it was solving a harder problem, with the caveat that the initial pose of the hand was correlated with the initial pose of the object, and we need to investigate in more detail to what extent the network can indeed perform the task blindly. Blind manipulation is of course not a requirement, given that many vision sensors now exist and real-time state estimators (Schmidt, Newcombe and Fox 2015) can provide object states. It will be interesting to test the network performance with more complete sensory input.

The nearest neighbour controller is currently time-varying: the switch happens at the beginning of the movement based on the initial state of the object, and then we use the selected local feedback controller for the rest of the movement. Another approach would be to store the dataset of states, controls and feedback gains along the local trajectories, and use nearest neighbour queries at each point in time. This requires more storage and processing than evaluating a neural network, but should not be a challenge for modern computing hardware. Furthermore, it may be possible to store a reduced set of controllers as opposed to storing the states along all trajectories. For example, Simbicon (Yin, Loken and van de Panne 2007) uses a small collection of time-invariant feedback controllers together with a suitable switching policy, and performs bipedal locomotion remarkably well. The Boston Dynamics approach to control, while not described in detail, also appears to be related.

It may be possible to combine the benefits of the two generalization methods considered here. While deep learning is currently popular, there is no reason to limit ourselves to generic networks. Instead we could consider a mixture-of-experts architecture (Jordan and Jacobs 1994), where the gating network corresponds to the switching mechanism in our current nearest neighbor approach, while the expert networks correspond to our local trajectory controllers. We would still want to leverage the power of trajectory optimization in training the experts, and perhaps use the current switching mechanism to pre-train the gating network.

Another direction for future work is to expand the set of tasks under consideration. While the present tasks are quite complex, they have been selected for their intrinsic stability. For example, consider our object-spinning task. If we were to attempt an identical task but with the palm facing down, our present approach would not work. This is because the object would drop before we have had time to interact with it and learn anything. In general, data-driven approaches to robotics require either an intrinsically stable task (which may be rare in practice), or a pre-existing controller that is able to collect relevant data without causing immediate failure. In this particular case we could perhaps use tele-operation to obtain such a controller, but a more general and automated solution is needed.

Finally, even though the focus of this paper is purely data-driven learning, there is no reason to take such a one-sided approach longer term. As with every other instance of learning from limited and noisy data, the best results are likely to be obtained when data is combined with suitable priors. In the case of robotics, physics provide a strong prior that can rule out the large majority of candidate models and control policies, thereby improving generalization from limited data on the physical system. Existing physics simulators can simulate complex systems such as the one studied here much faster than real-time, and can be run in parallel. This functionality can be used for model-predictive control, aided by a neural network representing a controller and/or a value function (Zhong, Johnson, Tassa, Erez and Todorov 2013). The model itself could be a hybrid between a physics-based model with a small number of parameters learned from data, and a neural network with a larger number of parameters used to fit the residuals that the physics-based model could not explain.

## Acknowledgements

This work was supported by the NIH, NSF and DARPA. The authors declare that there is no conflict of interest.

## References

Abbeel P, Coates A, Quigley M and Ng A (2006) An application of reinforcement learning to aerobatic helicopter flight. In: *Advances in Neural Information Processing Systems (NIPS)*.

Amend Jr JR, Brown E, Rodenberg N, Jaeger HM and Lipson H (2012) A positive pressure universal gripper based on the jamming of granular material. *Robotics, IEEE Transactions on* 28(2): 341–350.

Åström KJ and Wittenmark B (2013) *Adaptive control*. Courier Corporation.

Bagnell JA and Schneider J (2003) Covariant policy search. In: *International Joint Conference on Artificial Intelligence (IJCAI)*.

Bellman R and Kalaba R (1959) A mathematical theory of adaptive control processes. *Proceedings of the National Academy of Sciences* 8(8): 1288–1290.

Boyd S and Vandenberghe L (2004) *Convex Optimization*. New York, NY, USA: Cambridge University Press.

Deisenroth M, Neumann G and Peters J (2013) A survey on policy search for robotics. *Foundations and Trends in Robotics* 2(1-2): 1–142.

Deisenroth M, Rasmussen C and Fox D (2011) Learning to control a low-cost manipulator using data-efficient reinforcement learning. In: *Robotics: Science and Systems (RSS)*.

Gupta A, Eppner C, Levine S and Abbeel P (2016) Learning dexterous manipulation for a soft robotic hand from human demonstration. In: *IROS 2016*.

Jordan M and Jacobs R (1994) Hierarchical mixtures of experts and the em algorithm. *Neural Computation* 6(2): 181–214.

Kober J, Bagnell JA and Peters J (2013) Reinforcement learning in robotics: A survey. *International Journal of Robotic Research* 32(11): 1238–1274.

Kober J, Oztop E and Peters J (2010) Reinforcement learning to adjust robot movements to new situations. In: *Robotics: Science and Systems (RSS)*.

Kumar V and Todorov E (2015) Mujoco haptix: A virtual reality system for hand manipulation. In: *Humanoids*.

Kumar V, Todorov E and Levine S (2016) Optimal control with learned local models: Application to dexterous manipulation. In: *International Conference on Robotics and Automation (ICRA)*.

Kumar V, Xu Z and Todorov E (2013) Fast, strong and compliant pneumatic actuation for dexterous tendon-driven hands. In: *International Conference on Robotics and Automation (ICRA)*.

Levine S and Abbeel P (2014) Learning neural network policies with guided policy search under unknown dynamics. In: *Advances in Neural Information Processing Systems (NIPS)*.

Levine S, Finn C, Darrell T and Abbeel P (2015a) End-to-end training of deep visuomotor policies. *arXiv preprint arXiv:1504.00702* .

Levine S and Koltun V (2013) Guided policy search. In: *International Conference on Machine Learning (ICML)*.

Levine S, Wagener N and Abbeel P (2015b) Learning contact-rich manipulation skills with guided policy search. In: *International Conference on Robotics and Automation (ICRA)*.

Li W and Todorov E (2004) Iterative linear quadratic regulator design for nonlinear biological movement systems. In: *ICINCO (1)*. pp. 222–229.

Lioutikov R, Paraschos A, Neumann G and Peters J (2014) Sample-based information-theoretic stochastic optimal control. In: *International Conference on Robotics and Automation (ICRA)*.

Lynch K and Mason MT (1999) Dynamic nonprehensile manipulation: Controllability, planning and experiments. *International Journal of Robotics Research* 18(1): 64–92.

Mitrovic D, Klanke S and Vijayakumar S (2010) Adaptive optimal feedback control with learned internal dynamics models. In: *From Motor Learning to Interaction Learning in Robots*, volume 264. pp. 65–84.

Mnih V, Kavukcuoglu K, Silver D, Rusu AA, Veness J, Bellemare MG, Graves A, Riedmiller M, Fidjeland AK and Ostrovski G (2015) Human-level control through deep reinforcement learning. *Nature* 518(7540): 529–533.

Pastor P, Hoffmann H, Asfour T and Schaal S (2009) Learning and generalization of motor skills by learning from demonstration. In: *International Conference on Robotics and Automation (ICRA)*.

Peters J, Mülling K and Altun Y (2010a) Relative entropy policy search. In: *AAAI*. Atlanta.

Peters J, Mülling K and Altün Y (2010b) Relative entropy policy search. In: *AAAI Conference on Artificial Intelligence*.

Peters J and Schaal S (2008) Reinforcement learning of motor skills with policy gradients. *Neural Networks* 21(4).

Schmidt T, Newcombe R and Fox D (2015) Dart: dense articulated real-time tracking with consumer depth cameras. In: *Autonomous Robots*.

Sutton R and Barto A (1998) *Reinforcement Learning: An Introduction*. MIT Press.

Tedrake R, Zhang T and Seung H (2004) Stochastic policy gradient reinforcement learning on a simple 3d biped. In: *International Conference on Intelligent Robots and Systems (IROS)*.

Tesauro G (1994) Td-gammon, a self-teaching backgammon program, achieves master-level play. *Neural computation* 6(2): 215–219.

Theodorou E, Buchli J and Schaal S (2010) Reinforcement learning of motor skills in high dimensions. In: *International Conference on Robotics and Automation (ICRA)*.

Todorov E, Erez T and Tassa Y (2012) Mujoco: A physics engine for model-based control. In: *Intelligent Robots and Systems (IROS), 2012 IEEE/RSJ International Conference on*. IEEE, pp. 5026–5033.

van Hoof H, Hermans T, Neumann G and Peters J (2015) Learning robot in-hand manipulation with tactile features. In: *Humanoid Robots (Humanoids)*. IEEE.

Yin K, Loken K and van de Panne M (2007) Simbicon: Simple biped locomotion control. *ACM Trans. Graph.* 26(3): Article 105.

Zhong M, Johnson M, Tassa Y, Erez T and Todorov E (2013) Value function approximation and model-predictive control. In: *IEEE Symposium on Adaptive Dynamic Programming and Reinforcement Learning*.